\documentclass[letterpaper]{article} % DO NOT CHANGE THIS
\usepackage{aaai25}  % DO NOT CHANGE THIS
\usepackage{times}  % DO NOT CHANGE THIS
\usepackage{helvet}  % DO NOT CHANGE THIS
\usepackage{courier}  % DO NOT CHANGE THIS
\usepackage[hyphens]{url}  % DO NOT CHANGE THIS
\usepackage{graphicx} % DO NOT CHANGE THIS
\usepackage{natbib}  % DO NOT CHANGE THIS AND DO NOT ADD ANY OPTIONS TO IT
\usepackage{caption} % DO NOT CHANGE THIS AND DO NOT ADD ANY OPTIONS TO IT
\usepackage{algorithm}
\usepackage{algorithmic}
\usepackage{amsmath}
\usepackage{booktabs}
\usepackage{amsfonts}
\usepackage{amssymb}
\usepackage{subfigure}
\usepackage{caption}
\usepackage{subcaption}
\usepackage{multirow}
\usepackage{rotating}
\usepackage{newfloat}
\usepackage{listings}
\DeclareCaptionStyle{ruled}{labelfont=normalfont,labelsep=colon,strut=off} % DO NOT CHANGE THIS
\floatstyle{ruled}
\newfloat{listing}{tb}{lst}{}
\floatname{listing}{Listing}
\title{Two-Stage Representation Learning for Analyzing Movement Behavior Dynamics in People Living with Dementia}
\author{
Jin Cui\textsuperscript{\rm 1,2}\thanks{Corresponding author: \texttt{jc9223@ic.ac.uk}. Jin Cui and Alexander Capstick contributed equally.}, 
Alexander Capstick\textsuperscript{\rm 1,2}\footnotemark[1], 
Payam Barnaghi\textsuperscript{\rm 1,2}\thanks{Payam Barnaghi and Gregory Scott contributed equally.}, 
Gregory Scott\textsuperscript{\rm 1,2}\footnotemark[2]
}
\affiliations{
\textsuperscript{\rm 1}Imperial College London\\
\textsuperscript{\rm 2}UK Dementia Research Institute, Care Research and Technology Centre
}

\begin{document}

\maketitle

\begin{abstract}
  In remote healthcare monitoring, time series representation learning reveals critical patient behavior patterns from high-frequency data. This study analyzes home activity data from individuals living with dementia by proposing a two-stage, self-supervised learning approach tailored to uncover low-rank structures. The first stage converts time-series activities into text sequences encoded by a pre-trained language model, providing a rich, high-dimensional representation. In the second stage, these vectors are transformed into a low-dimensional latent state space using a PageRank-based method. This PageRank vector captures latent state transitions, effectively compressing complex behavior data into a succinct form that enhances model interpretability. This low-rank representation not only enhances interpretability but also facilitates clustering and transition analysis, revealing key behavioral patterns correlated with clinical metrics such as MMSE and ADAS-Cog scores. Our findings demonstrate the framework’s potential in supporting cognitive status prediction, personalized care interventions, and large-scale health monitoring. 
\end{abstract}

\section{Introduction}

\begin{figure}[htbp]
    \centering
    \includegraphics[width=0.4\textwidth]{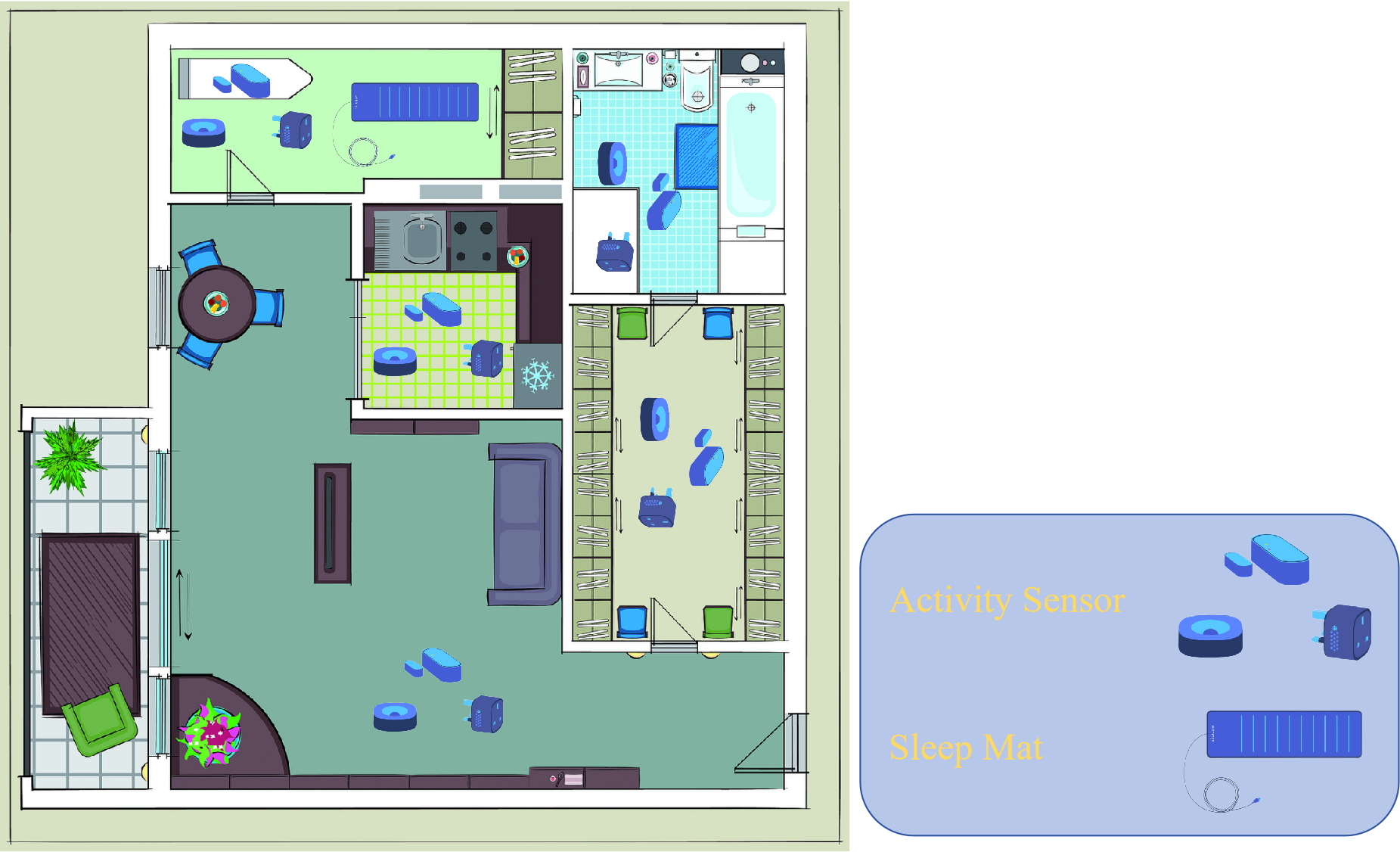}
    \caption{Example Home layout with IOT sensors for monitoring behavioral patterns of People Living with Dementia}
    \label{fig:minder_home}
\end{figure}

In remote healthcare monitoring applications, the use of wearables and Internet of Things (IoT) devices to continuously collect time-series data, as is shown in Figure \ref{fig:minder_home}, often with second-level accuracy or finer, has become increasingly common. However, the sheer scale of such data makes it difficult for human experts to analyze or use directly, necessitating the use of time-series deep learning techniques for effective analysis and diagnosis.

Training on large volumes of unlabeled time-series data poses a significant challenge. Semi-supervised and unsupervised methods are typically employed to encode and extract data features for downstream tasks like classification or regression, demonstrating their ability to capture deep features. Semi-supervised methods, such as nearest neighbor contrastive learning and temporal relation prediction, efficiently utilize both labeled and unlabeled data, improving the quality of representations for downstream tasks like classification \citep{kim_semi-supervised_2024,fan_semi-supervised_2021}. Unsupervised methods focus on learning robust representations without relying on labels, often leveraging contrastive learning techniques and innovative data augmentations to capture key temporal patterns \citep{franceschi_unsupervised_2019,lee_spatio-temporal_2024}. Attention mechanisms and domain-adaptive techniques further enhance the interpretability of encoded features, aligning them more closely with human intuition and domain-specific insights \citep{lyu_improving_2018}. However, this strategy faces two challenges: first, labeling criteria for time-series data is often vague, which can significantly impact model performance; second, the encoded data remain vast, unintuitive, and difficult to interpret \citep{ye_lbp4mts_2023,hill_semi-supervised_2022}.

In this work, we focus on time-series data characterized by irregular discrete values. Extending the methods introduced in \citep{capstick2024representation}, we present preliminary results of a second-order representation learning method designed to aid in clustering, identifying similar clinical cases, and uncover patients' interpretable behavioral patterns. This is achieved through a large language model encoding combined with a two-dimensional vectors representation and transfer pattern analysis.

\subsection{Our Contribution}

We propose an integrated approach for discovering latent states of activity. This method comprises several key steps:

\begin{enumerate}
    \item \textbf{Temporal Data Preprocessing}: The raw temporal data is first preprocessed to remove noise and standardize the data for consistency.
    \item \textbf{Language Model Encoding}: A language model is trained on our dataset to encode the preprocessed temporal data into high-dimensional vector representations. To enhance the model's learning capability, we perform pseudo labeling using one-hot similarity. This allows the model to better capture temporal dependencies and patterns in the data.
    \item \textbf{Dimensionality Reduction and Clustering}: To visualize the high-dimensional embeddings, we apply dimensionality reduction techniques such as t-SNE to project the data into a 2D space. Clustering algorithms are then used to identify distinct latent states within the data.
    \item \textbf{Transition Pattern Analysis with PageRank}: By defining a transition matrix between these low-rank latent states, we use the PageRank algorithm to analyze transition patterns. This approach compresses complex temporal data into interpretable, low-rank state vectors, allowing us to determine the influence and significance of each state within the transition graph and provide insights into patient behavior dynamics.
\end{enumerate}

This analytical framework will aid in the clinical diagnosis of patients and support the development of personalized care programs. The availability of dataset and code for this work is discussed in appendix.

\section{Related Work}

Time-series forecasting is primarily to predict future values based on previously observed data points. Traditional statistical methods, most notably the Autoregressive Integrated Moving Average (ARIMA) model, have long been utilized due to their mathematical simplicity and flexibility in application \citep{rizvi_arima_2024,kontopoulou_review_2023}. While ARIMA remains a staple for scenarios where data exhibits linear patterns, recent developments in machine learning have introduced sophisticated models capable of capturing non-linear dependencies, thus offering potential improvements in forecasting accuracy and robustness \citep{masini_machine_2023,rhanoui_forecasting_2019}.

The advent of the Generative Pre-trained Transformer (GPT) by OpenAI marked a significant milestone in the field of natural language processing \citep{brown_language_2020}, catalyzing a wave of innovations in large language models (LLMs). Large Language Models (LLMs) have profoundly transformed natural language processing and are increasingly being considered for diverse applications beyond text, such as time series data analysis. The study by \citep{bian_multi-patch_2024} presents a framework that adapts LLMs for time-series representation learning by conceiving time-series forecasting as a multi-patch prediction task, introducing a patch-wise decoding layer that enhances temporal sequence learning. Similarly, \citep{liu_autotimes_2024} propose a model which leverages the autoregressive capabilities of LLMs for time series forecasting. In \citet{capstick2024representation}, the authors apply a GPT-based text encoder to string representations of in-home activity data to enable vector searching and clustering. Using a secondary modelling stage, we extend these ideas to enable further analysis and interpretability.

PageRank, originally developed to rank web pages, is an algorithm designed to assess the importance of nodes within a directed graph by analyzing the structure of links within networks \citep{page_pagerank_1999}. While it was initially created for search engines, its application has since expanded across various disciplines. For instance, in biological networks, \citep{ivan_when_2011} employed personalized PageRank to analyze protein interaction networks, providing scalable and robust techniques for interpreting complex biological data. Similarly, \citep{banky_equal_2013} introduced an innovative adaptation of PageRank for metabolic graphs. This cross-disciplinary application of PageRank highlights its potential for analyzing complex systems beyond its original domain.

\section{Methods}
\label{method}

\subsection{Mathematical Foundations of the Model}
Given a discrete data sample $\mathbf{X} = \{x_1, x_2, \dots, x_n\}$, the following steps describe the transformation process:

1. \textbf{Sampling and Text Conversion:} Each sample $x_i$ is converted into a text representation $T(x_i)$.

2. \textbf{Language Model Encoding:} A pre-trained language model $f_{\text{LM}}$ is applied to obtain high-dimensional vector embeddings for the text data:
\[
\mathbf{h}_i = f_{\text{LM}}(T(x_i)), \quad \mathbf{h}_i \in \mathbb{R}^d.
\]

3. \textbf{Dimensionality Reduction:} The high-dimensional embeddings are projected into a 2D space using a dimensionality reduction method $\Phi$, such as t-SNE:
\[
\mathbf{z}_i = \Phi(\mathbf{h}_i), \quad \mathbf{z}_i \in \mathbb{R}^2.
\]

4. \textbf{PageRank and Deep State Vector Extraction:} A transition matrix $\mathbf{P}$ between points in 2D space is constructed, and the PageRank algorithm is applied to further reduce the dimensionality:
\[
\mathbf{v}_i = \text{PageRank}(\mathbf{P}), \quad \mathbf{v}_i \in \mathbb{R}^k, \quad k \ll d.
\]

The final low-dimensional vectors $\mathbf{v}_i$ capture deep semantic relationships from the original data.

\subsection{The Dataset}

We obtained a dataset collected from 134 people diagnosed with dementia, capturing their home location movement data between July 1, 2021, and January 30, 2024. The dataset records the time entering different rooms and sleeping mats, alongside clinical metrics such as MMSE \citep{kurlowicz_mini-mental_1999}, ADAS-Cog \citep{kueper_alzheimers_nodate} scores from regular tests. It also includes details on various factors such as demographic data, comorbidities, and other medical information. The dataset contains a total of 66,096 recording days. A more detailed description of the dataset is provided in Appendix. After excluding patients with missing data, the final dataset used for further analysis contained 50 participants with complete information.

\subsection{Our Framework}

Our framework consists of several key stages: data preprocessing and encoding, latent state discovery, and transition pattern analysis. 

\begin{figure*}[htbp]
    \centering
    \includegraphics[width=0.8\textwidth]{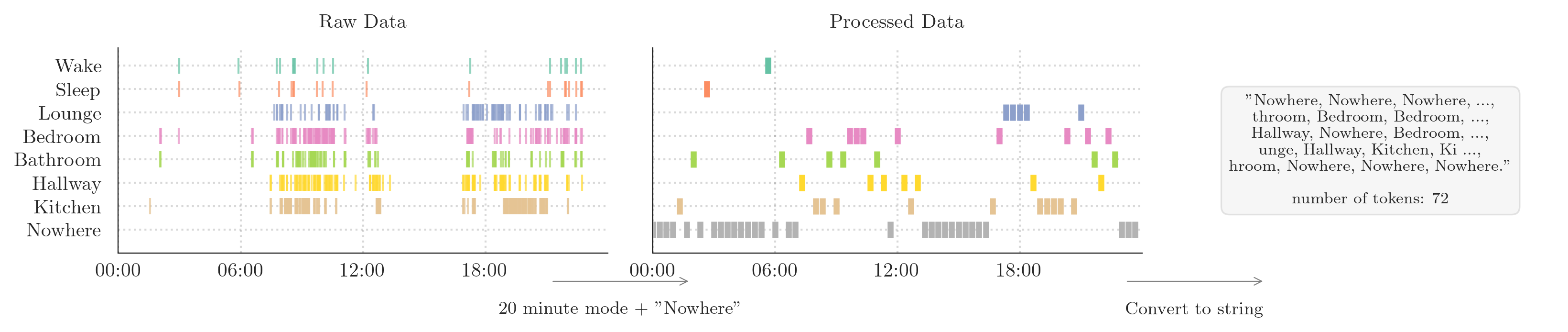}
    \caption{Flowchart of data preprocessing. The figure illustrates the monitoring data for a single participant over the course of one day. The left graph displays the raw, unprocessed measurements. In the middle graph, the data is rectified into 20-minute intervals, where periods of inactivity are labeled as "nowhere." Within each window, the most frequent location, excluding "nowhere," is identified and recorded. The right graph presents the corresponding text strings, which are formatted for interpretation by the language model.}
    \label{fig:preprocess-diagram}
\end{figure*}

First, we preprocess the raw temporal data to remove noise and ensure consistency. This process is illustrated in Figure \ref{fig:preprocess-diagram}. We then utilize the all-MiniLM-L12-v2 model \citep{muennighoff_mteb_2023} as the language model encoder. This model excels at capturing similarities in textual information, making it suitable for analyzing similarities between recorded dates and uncovering potential relationships. We fine-tune the model using its pretrained weights to adapt to the specific characteristics of our dataset. The preprocessed temporal data is then encoded into 384-dimensional vector representations, capturing the inherent temporal dependencies and patterns within the data.

\begin{figure}[htbp]
    \centering
    \includegraphics[width=0.4\textwidth]{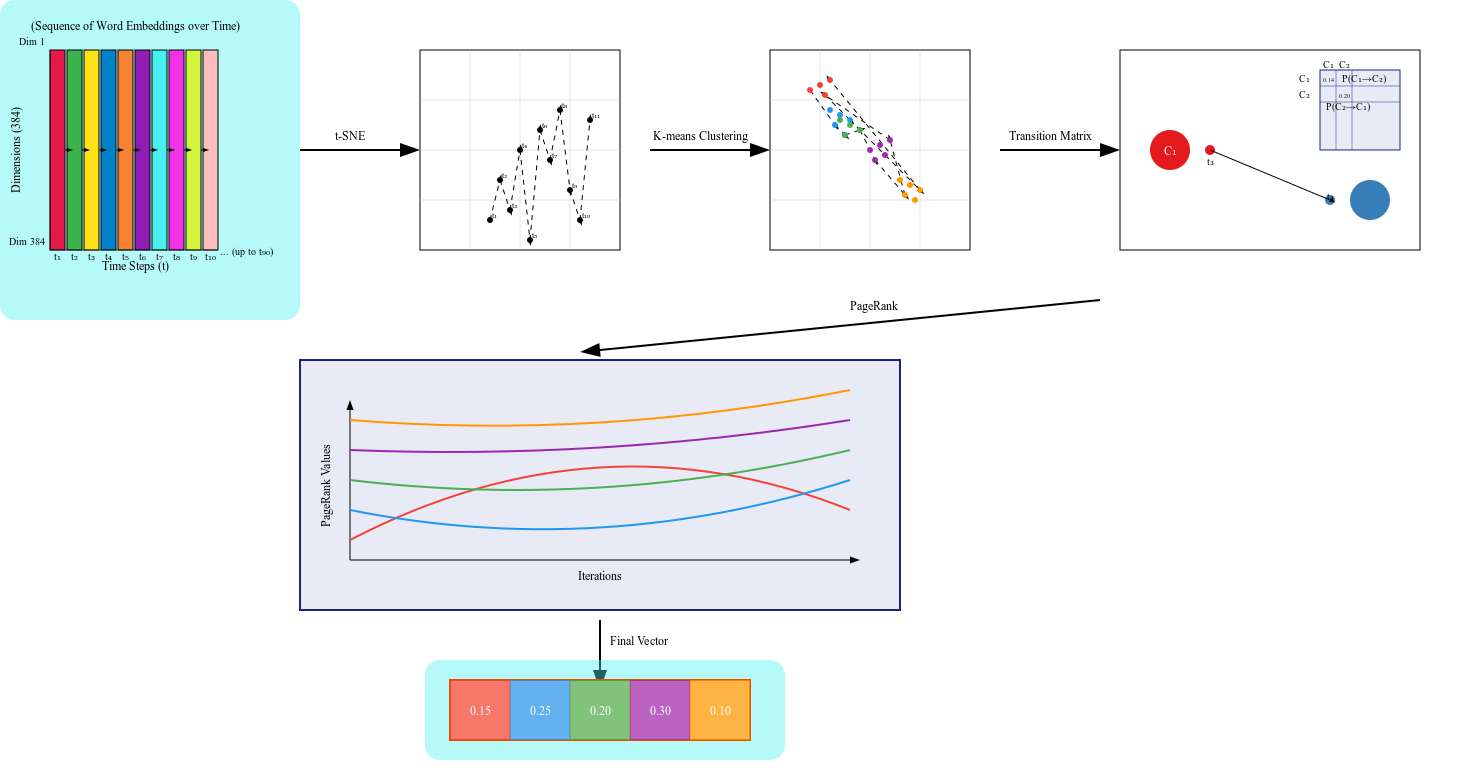}
    \caption{Flowchart of the representation algorithm.}
    \label{fig:flowchart-diagram}
\end{figure}

 Given the unlabeled nature of our temporal data, we adopt a cluster-based contrast sample selection method and triplet loss for training and evaluation. Further details on the language model training process are described in Appendix. To visualize and interpret the high-dimensional embeddings, we apply the t-SNE dimensionality reduction technique \citep{van_der_maaten_viualizing_2008} to project the data into a 2D space. K-means clustering is then employed to identify distinct latent states within the data. As the data points are temporally ordered, this 2D map allows us to visualize each participant's latent activity map as their movement pattern projected onto a specific dimension. Finally, we define a transition matrix between the different latent states and apply the PageRank algorithm \citep{page_pagerank_1999} to analyze the transition patterns, as is shown in Figure \ref{fig:flowchart-diagram}, details of this algorithm are available in Appendix. In this study, we present a scalable approach to analyze vast amounts of temporal data in patient movement behavior. Consider a single participant with continuous, complete data collected over a three-month period. Given that the passive infrared (PIR) sensors record data in seconds, the approximate dataset size would be 3 × 30 × 86,400 = 7,776,000 sparse records for a single individual. Such a data volume is infeasible for direct analysis by human experts and would require extensive computational resources to process with most machine learning models. Our method addresses this challenge by transforming this high-dimensional dataset into a low-dimensional, semantically interpretable representation. Specifically, through our two-stage encoding and PageRank-based dimensionality reduction, we compress the data into a latent state vector of length five. Each element in this vector is not only computationally efficient but also carries interpretable semantic information, facilitating transparent and personalized insights into patient behavior dynamics. This approach exemplifies the application of low-rank structures to manage and interpret complex temporal datasets in healthcare AI, supporting personalized intervention strategies.

\section{Experiments}
\label{experiment}
\begin{figure}[htbp]
    \centering
    \includegraphics[width=0.4\textwidth]{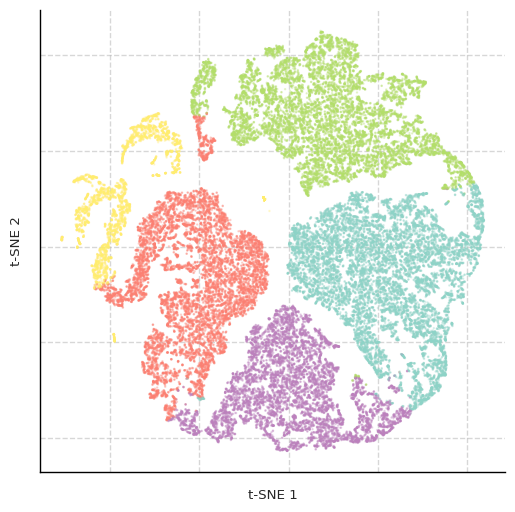}
    \caption{T-SNE for embedded daily movement strings in the test set}
    \label{tsneplotpic1}
\end{figure}

After clustering the text vectors of the test set using K-means, we identified the optimal clustering result at 5 clusters, suggesting five latent states across all single-day, single-participant behavioral patterns. Figure \ref{tsneplotpic1} shows the clustering results after the dimension reduction of the embeddings using t-SNE. By examining the transformation of individual vectors in two dimensions, we can visualize the behavioral trajectories of different participants within the embedding space (see appendix for more participant visualizations). Collaborating with clinical experts, we can explore the semantics represented by these clusters and their relationship to patient medical characteristics.

\begin{figure}[htbp]
    \centering
    \includegraphics[width=0.4\textwidth]{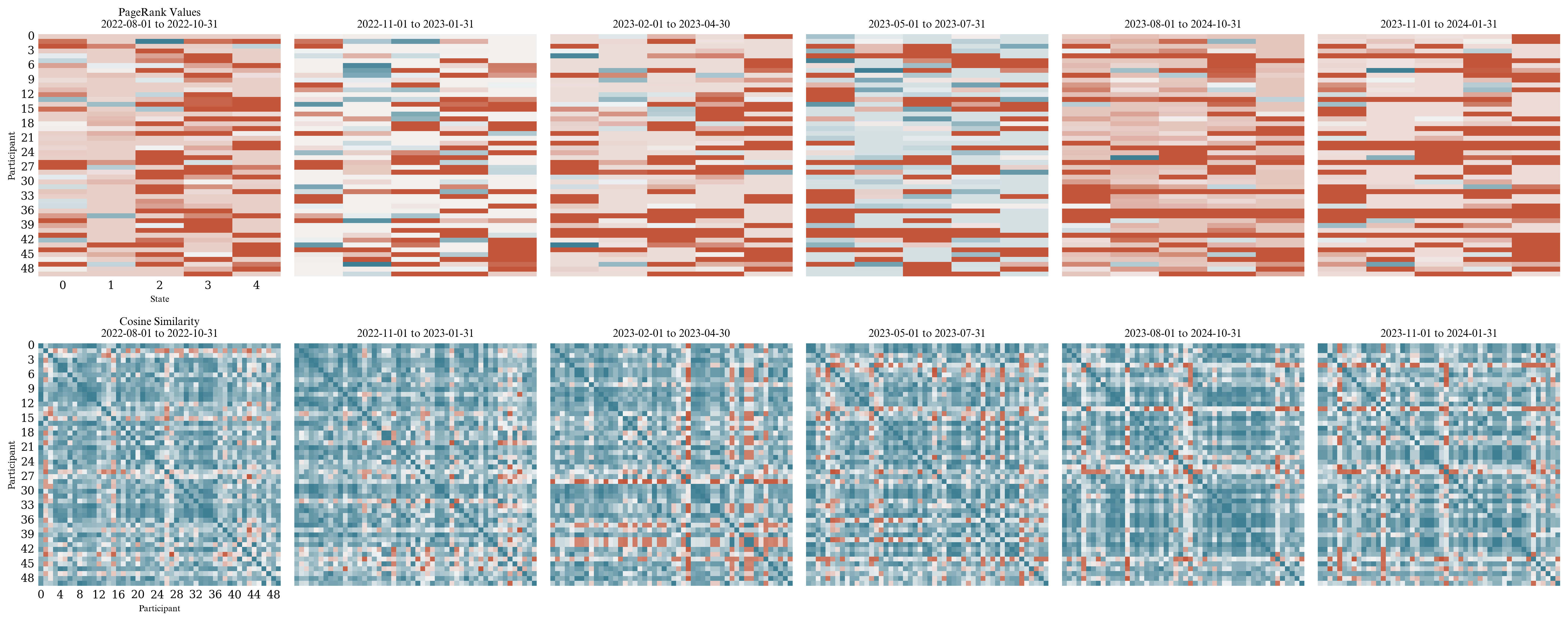}
    \caption{Multi-period participant deep state vector and similarity}
    \label{multiperiod}
\end{figure}

More significantly, by applying the random walk model and the PageRank algorithm to these two-dimensional plots, in combination with clinical expert opinions and diagnostic results, we can quantitatively assess the deeper semantics represented by the vector clusters, or latent states. Figure \ref{multiperiod} presents a multi-period heatmap analysis, segmented across five time intervals to visualize patient behavior dynamics. The top row of heatmaps illustrates the PageRank values in different latent states (1-5) for each participant over time, with each interval representing a period of three months. The intensity of color in each cell reflects the PageRank value for a particular state and participant, highlighting shifts in dominant latent states and revealing periodic patterns in behavior. This aggregation reduces the high-dimensional behavioral data to a low-rank, interpretable representation. The bottom row of the heatmaps shows the cosine similarity matrices between the participants for each respective time interval. These matrices indicate how similar participants' behaviors are to each other within each period, with higher similarity values shown in darker shades. The block patterns visible in some intervals suggest clustering tendencies among participants with similar behavioral patterns. The heatmap analysis reveals distinct behavioral patterns among participants, with certain individuals demonstrating notable stability and periodicity in their daily activity states. For example, several participants consistently show high PageRank values for specific latent states (such as state 2 or 3) across multiple time intervals. This stability may indicate regular lifestyle patterns or consistent routines, which can be critical for predicting daily behavior and planning personalized interventions. In contrast, some participants exhibit significant shifts in their PageRank values between periods, particularly around seasonal transitions or specific time frames (e.g., August-October 2022 and May-July 2023). These variations may reflect seasonal effects or environmental changes influencing patient behavior, underscoring the need to account for time-related factors when designing personalized monitoring and intervention strategies. The cosine similarity heatmaps reveal clusters of participants who display high similarity in behavior during certain intervals (e.g., May–July 2023 and August 2023–January 2024), forming identifiable groups. This similarity may suggest shared behavioral characteristics or similar health conditions between these participants, possibly due to common lifestyle factors, environmental influences, or comparable stages in disease progression. These group-level patterns provide information on cohort-based behavioral dynamics, supporting the development of individualized and group-based healthcare management strategies.

Based on the clustering and SHAP analyses of PageRank-derived states and cognitive metrics, as is shown in appendix, distinct behavioral and cognitive profiles emerge across the clusters. Each PageRank state, as visualized through correlation and SHAP values, highlights unique relationships with cognitive scores (MMSE, ADAS-Cog), age, and mood-related metrics (HADS Depression and Anxiety). In the appendix, by reducing the complex vector matrix to a simplified (1,5) vector, we can explore semantic characteristics to each of the states.

Looking forward, once a unique deep vector is generated for each participant, we can use these states to predict MMSE, ADAS-Cog scores, and their rate of change. Detailed methodologies and implementation specifics are provided in the appendix for further reference. 

\subsection{Cognitive Status Prediction Performance}

To evaluate the impact of generated state features on predictive performance, we consider several feature combinations, including:
\begin{itemize}
    \item \textbf{Baseline:} Original location count-based features (mean and variance).
    \item \textbf{State Features:} Representation of patient states (\texttt{state1} to \texttt{state5}).
    \item \textbf{Characteristics:} Patient-specific characteristics such as age, gender, and dementia diagnosis.
    \item \textbf{Combined Features:} Various combinations of the above, including all features.
\end{itemize}

\begin{table*}[htbp]
\caption{Performance metrics for predicting current ADAS-Cog and MMSE with 95\% confidence intervals (CI). The values in parentheses represent the lower and upper bounds of the 95\% CI. }
\resizebox{\textwidth}{!}{
\begin{tabular}{llllll}
\toprule
 & \textbf{Metric} & \textbf{MAE\textsubscript{ADASCOG}} & \textbf{MAE\textsubscript{MMSE}} & \textbf{RMSE\textsubscript{ADASCOG}} & \textbf{RMSE\textsubscript{MMSE}} \\
\textbf{Model} & \textbf{Feature Set} &  &  &  &  \\
\midrule
\multirow[t]{6}{*}{RandomForest} & Baseline & 11.41 (9.20, 13.65) & 4.64 (3.83, 5.50) & 11.37 (9.25, 13.64) & 4.68 (3.90, 5.50) \\
 & Baseline + Characteristics & 10.81 (8.63, 13.10) & 4.55 (3.76, 5.43) & 10.85 (8.76, 13.24) & 4.59 (3.73, 5.51) \\
 & Characteristics & 10.62 (8.27, 13.08) & 4.34 (3.39, 5.36) & 10.65 (8.24, 13.13) & 4.33 (3.34, 5.30) \\
 & State & 10.46 (8.31, 12.63) & 4.07 (3.09, 5.03) & 10.41 (8.25, 12.54) & 4.09 (3.20, 5.09) \\
 & State + Baseline + Characteristics & 10.50 (8.43, 12.69) & 4.37 (3.58, 5.25) & 10.62 (8.39, 12.84) & 4.39 (3.50, 5.30) \\
 & State + Characteristics & 10.31 (8.21, 12.79) & 4.19 (3.32, 5.06) & 10.34 (8.33, 12.55) & 4.19 (3.30, 5.13) \\
\midrule
\multirow[t]{6}{*}{Ridge} & Baseline & 12.77 (10.23, 15.39) & 5.15 (4.11, 6.31) & 12.71 (10.24, 15.22) & 5.10 (4.00, 6.30) \\
 & Baseline + Characteristics & 12.04 (9.67, 14.43) & 5.03 (4.01, 6.17) & 11.93 (9.66, 14.34) & 5.05 (4.08, 6.16) \\
 & Characteristics & 10.34 (8.33, 12.47) & 4.35 (3.51, 5.34) & 10.31 (8.25, 12.53) & 4.34 (3.47, 5.37) \\
 & State & \textbf{9.73} (7.31, 12.20) & \textbf{3.81} (2.81, 4.89) & \textbf{9.83} (7.38, 12.42) & \textbf{3.81} (2.74, 4.93) \\
 & State + Baseline + Characteristics & 13.20 (10.53, 16.14) & 5.49 (4.40, 6.53) & 13.23 (10.73, 16.01) & 5.49 (4.39, 6.59) \\
 & State + Characteristics & 10.18 (7.84, 12.87) & 4.27 (3.26, 5.32) & 10.15 (7.69, 12.70) & 4.25 (3.18, 5.42) \\
\bottomrule
\end{tabular}}
\label{tab:current_table}
\end{table*}

Based on the performance metrics presented in the Table \ref{tab:current_table}, we evaluated the effectiveness of different feature sets in predicting current ADAS-Cog and MMSE scores. Details of the specific model parameters, training configurations, and feature set explanation can be found in the appendix.

First, the baseline model, which only used the mean and variance of participants' daily activity data as features, provided some predictive capability, but with relatively large margins of error. In the Random Forest model, the baseline features achieved an MAE\textsubscript{ADASCOG} of 11.41 and an MAE\textsubscript{MMSE} of 4.64, while in the Ridge regression model, these values were 12.77 and 5.15, respectively, indicating substantial prediction errors under this feature set.

When combining baseline features with patient clinical characteristics (such as HADS scores and age), the model's performance improved modestly. However, further analysis revealed that using the deep state features alone resulted in significant performance improvements. In the Ridge regression model, the state features achieved an MAE\textsubscript{ADASCOG} of 9.73 and an MAE\textsubscript{MMSE} of 3.81, which were the best results across all feature combinations. The corresponding RMSE values were also low, at 9.83 and 3.81, respectively. Compared to the baseline features or combined feature sets, the state features demonstrated higher predictive accuracy and narrower 95% confidence intervals.

The Random Forest model also showed relatively strong performance when using state features, with an MAE\textsubscript{ADASCOG} of 10.46 and an MAE\textsubscript{MMSE} of 4.07. Although slightly less accurate than the Ridge model, it still showed improvement over the baseline model.

Analysis of the combined feature sets further underscores the strong capacity of the deep state features in capturing patient behavior patterns and cognitive states. Notably, the Ridge regression model achieved the best performance when using only the state features, suggesting that these compact deep vectors effectively capture patient cognitive patterns. Compared to conventional baseline features and patient demographic information, the state features provide a more powerful representation, offering a promising avenue for future clinical applications.

Now that we have established a process for encoding deep vectors, we could explore transforming this approach into a generative model. Such a model could be used to generate sensitive and hard-to-obtain medical datasets for purposes like data augmentation or alignment, a strategy proven effective in the training of large language models\citep{li_synthetic_2023}.

\section{Conclusion}

In conclusion, our initial results demonstrate that by applying our framework, we show that our latent states vector based on patient daily activity patterns can be useful for exploring behavior dynamics. While these findings offer a promising approach to exploring the relationship between behavior and clinical characteristics, further research is needed to refine the model and validate its broader applications, including potential use in medical data augmentation.

\bibliography{Formatting-Instructions-LaTeX-2025}

%%%%%%%%%%%%%%%%%%%%%%%%%%%%%%%%%%%%%%%%%%%%%%%%%%%%%%%%%%%%

%%%%%%%%%%%%%%%%%%%%%%%%%%%%%%%%%%%%%%%%%%%%%%%%%%%%%%%%%%%%

\appendix
\appendix
\section{Appendix}

\subsection{Model Pipeline}
\begin{figure*}[htbp]
    \centering
    \includegraphics[width=0.8\textwidth]{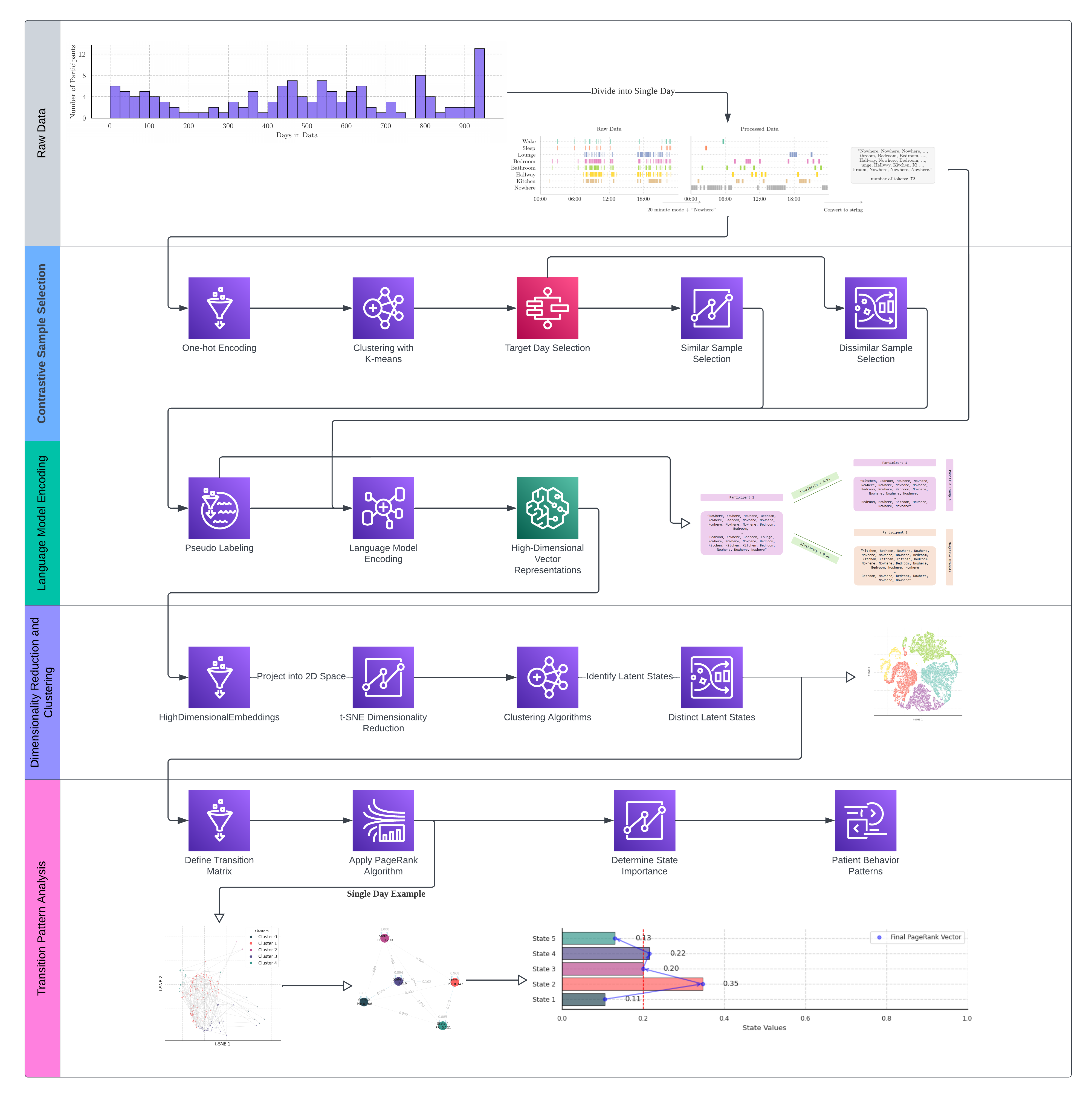}
    \caption{Model pipeline}
    \label{pipeline}
\end{figure*}

As is shown in Figure \ref{pipeline}, the workflow for analyzing the dynamics of the patient's daily movement behavior dynamics involves five primary stages. First, raw temporal data is preprocessed to remove noise and segment multiple-day records into single-day datasets. In the Contrastive Sample Selection phase, one-hot encoding and K-means clustering are applied to create clusters, facilitating the selection of similar and dissimilar samples for contrastive learning. Next, Language Model Encoding utilizes a language model to generate high-dimensional vector representations, with pseudo-labeling enhancing temporal dependency capture. The Dimensionality Reduction and Clustering stage uses t-SNE to project embeddings into a 2D space, allowing clustering algorithms to identify distinct latent states. Finally, Transition Pattern Analysis defines a transition matrix across latent states and applies the PageRank algorithm to evaluate state importance, yielding a compressed representation of patient behavior patterns that enhances interpretability and supports personalized intervention strategies.

\subsection{Availability of Datasets and Code}
\label{append_data_reproduction}

The IPython notebooks used to build this framework will be released after review. The dataset and IPython notebooks used to plot the data will not be made public due to their sensitivity. The experiments were conducted using Python 3.11.5, Torch 2.4.0 \citep{ansel_pytorch_2024}, Transformers 4.44.2 \citep{wolf_transformers_2020}, Sentence-Transformers 2.7.0 \citep{pedregosa_scikit-learn_2011}, Scikit-Learn 1.3.2 \citep{pedregosa_scikit-learn_2011}, NumPy 1.26.4 \citep{harris_array_2020}, SciPy 1.13.1 and Pandas 2.1.4 \citep{mckinney_data_2010}. 

\subsection{Detailed Description of The Dataset}
\label{append_dataset}

\begin{figure}[htbp]
    \centering
    \includegraphics[width=0.4\textwidth]{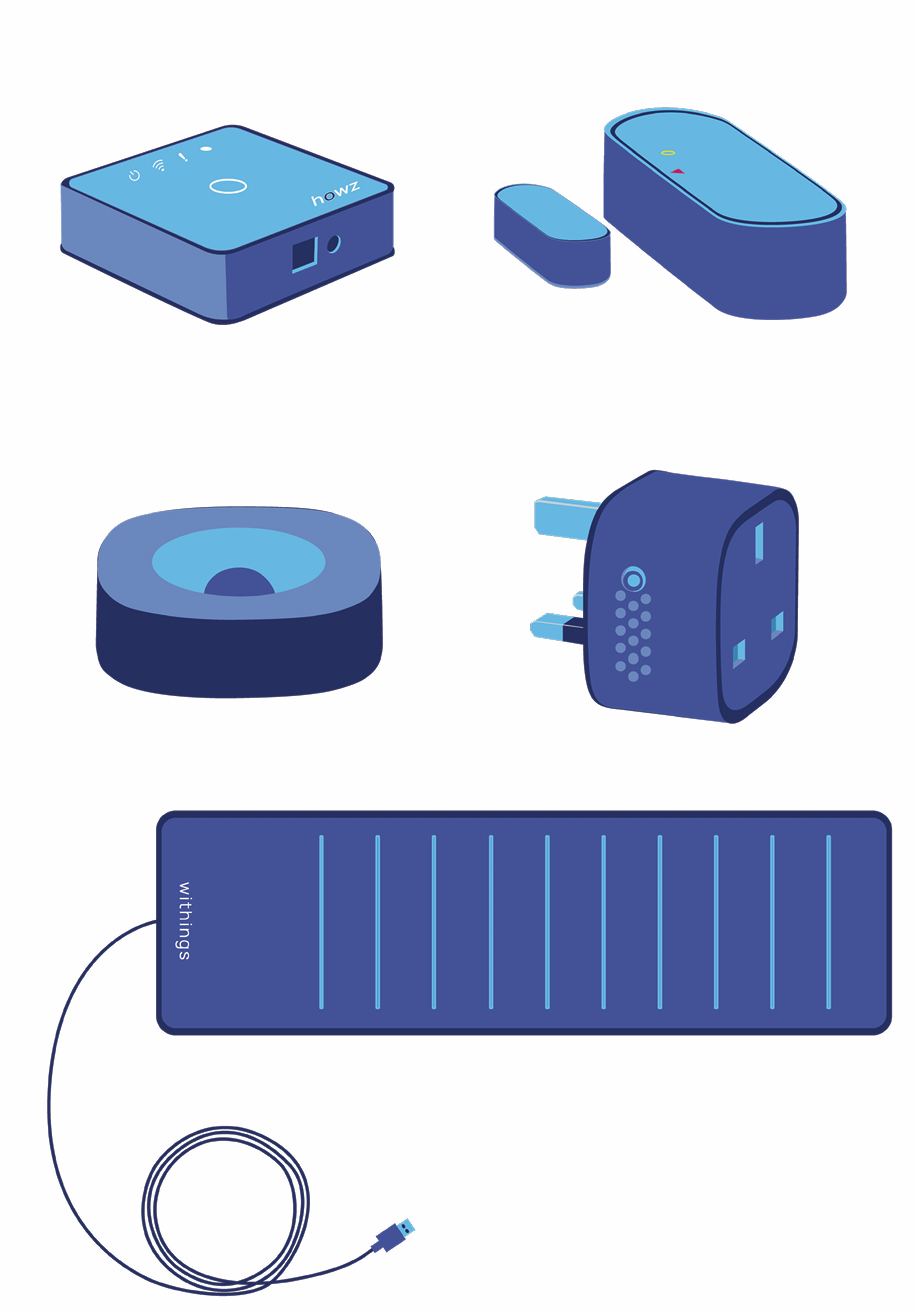}
    \caption{Minder sensors}
    \label{minder}
\end{figure}

The dataset used for in-home activity monitoring was collected via passive infrared sensors installed at multiple locations in the homes of individuals with dementia, along with sleep pads placed under their mattresses, as is shown in Figure \ref{minder}, from Minder website. These infrared sensors detect motion within a range of up to nine meters, at a maximum angle of forty-five degrees diagonally upward. Sensors were placed in lounges, kitchens, hallways, bedrooms, and bathrooms, allowing for detailed tracking of participants' movements within and between these areas.

\begin{figure}[htbp]
    \centering
    \includegraphics[width=0.4\textwidth]{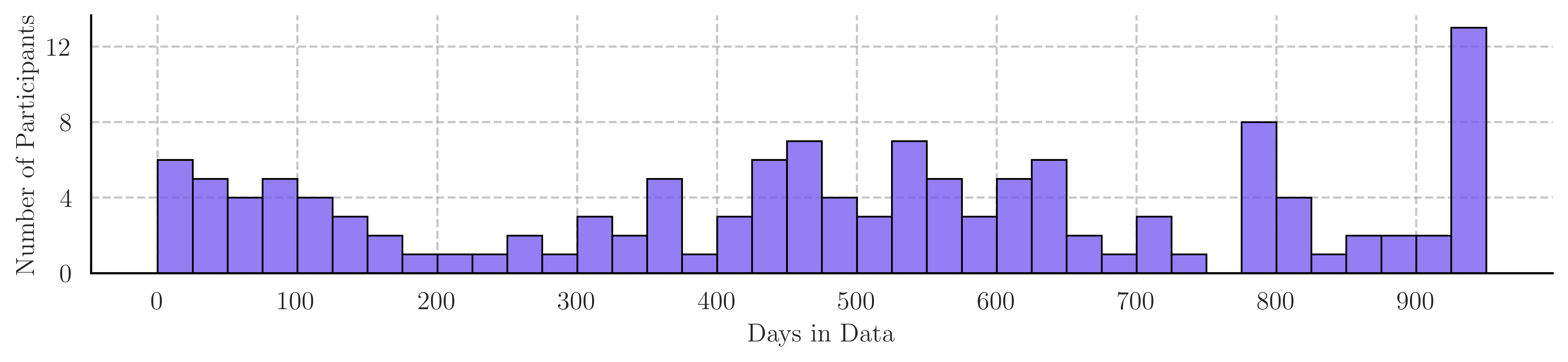}
    \caption{Daily histogram}
    \label{histo}
\end{figure}

\begin{figure}[htbp]
    \centering
    \includegraphics[width=0.4\textwidth]{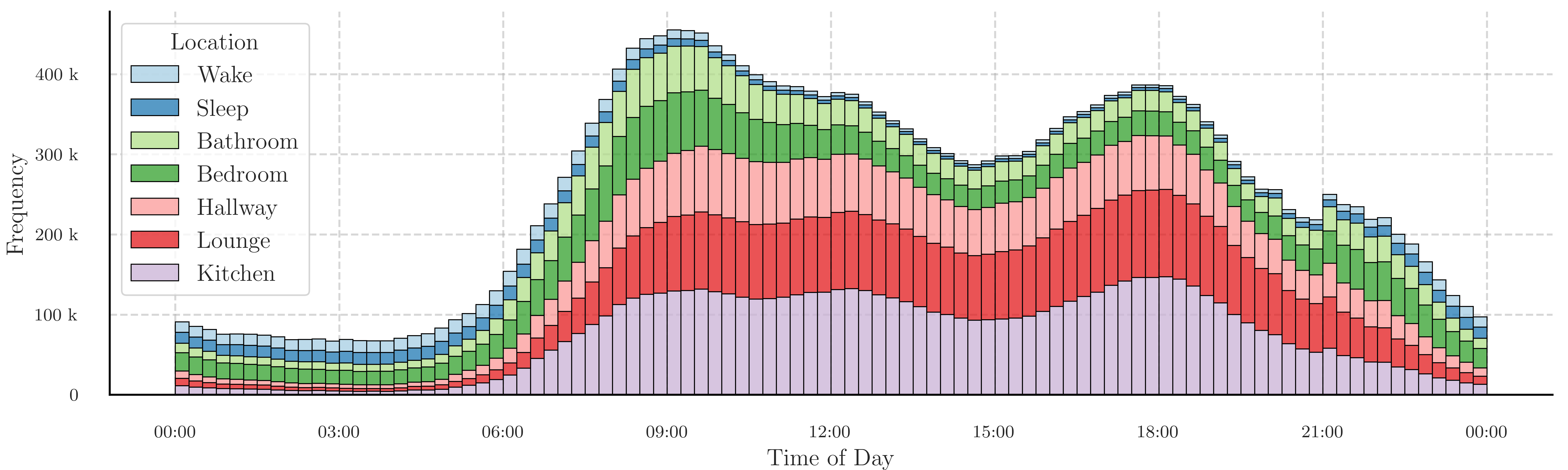}
    \caption{Location histogram}
\label{location}
\end{figure}

\begin{figure}[htbp]
    \centering
    \includegraphics[width=0.4\textwidth]{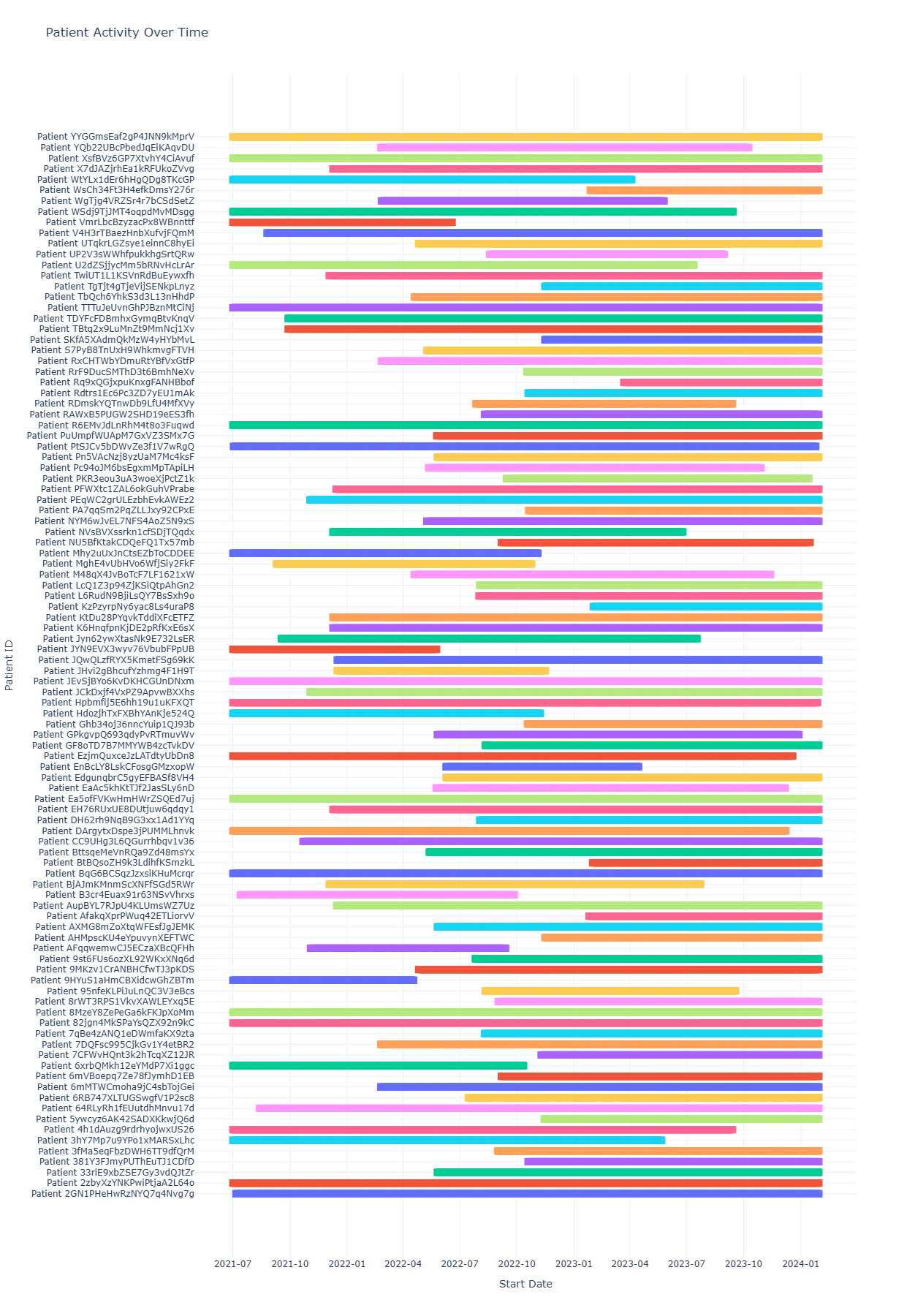}
    \caption{Top 50 timeseries distribution}
\label{threemonth}
\end{figure}

We analyzed data recorded between July 1, 2021, and January 30, 2024, amounting to 66,096 participant-days for 134 individuals. Figure \ref{histo} illustrates the distribution of logged days per participant. Each data point includes the participant ID, a timestamp (accurate to the second), the location of detected activity, or sleep pad data indicating whether the participant entered or left their bed. The dataset contains a total of 24,467,307 individual records, as depicted in Figure \ref{location}. Figure \ref{threemonth} shows the top 50 patients with the longest recorded lengths of time-span.

\begin{figure}[htbp]
    \centering
    \includegraphics[width=0.4\textwidth]{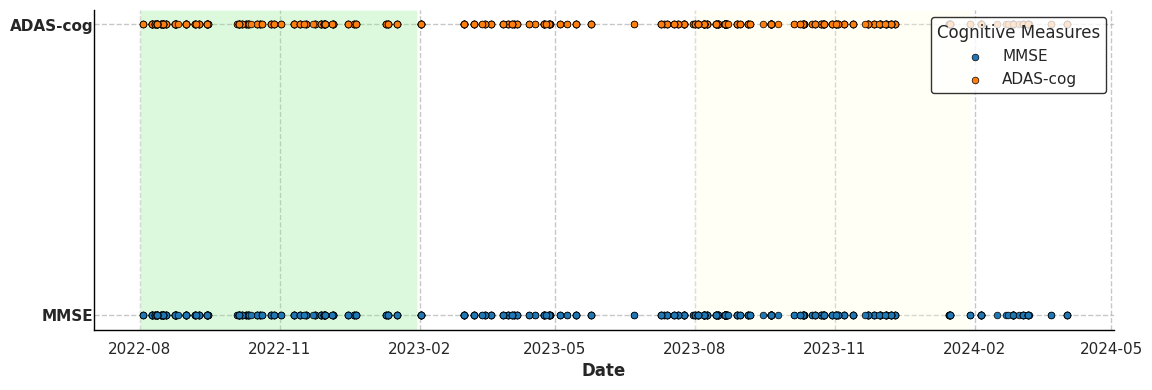}
    \caption{Timeseries of cognitive test of the test set participants}
    \label{cognitive}
\end{figure}

In addition to activity data, we had access to diagnostic information for the 134 participants, including birthdate, gender, living situation (whether they lived alone), ethnicity, and dementia diagnosis. Cognitive assessment scores, such as MMSE and ADAS-Cog, along with their yearly changes, were also available, as is shown in Figure \ref{cognitive}. To mitigate the risk of time series data contamination, the period from July 1, 2021, to July 1, 2023, was used for fine-tuning, while data from July 31, 2023, to January 30, 2024, was reserved for testing. The test set data was used to directly extract coding vectors from the fine-tuned language model and analyze behavioral transfer patterns. However, participants with incomplete records, particularly those with gaps in data after July 2023, were excluded from the test set. The final test set comprised the fifty participants with the most complete data post-July 2023, resulting in 869 comprehensive clinical records. The cognitive test results of participants in the test set, from August 2023 through the cut-off date of January 30, 2024, were used for our analyses. Additionally, we incorporated their cognitive test results from one year earlier to assess changes in MMSE and ADAS-Cog scores over time.

\subsection{Detailed Training Process}
\label{append_training}

Since the sensor data are recorded with second-level precision, each participant generates 86,400 data points per day, far exceeding the input token limit of the language model we are using (which allows a maximum of 256 tokens). To address this, the raw data were downsampled by extracting discrete values at 20-minute intervals, reducing the data points per day to 72. After converting these data points to strings, the token count is 72, which falls within the model's token limit.

Given the unlabeled nature of our temporal data, we employ a cluster-based contrastive sample selection approach for model training. This method leverages the inherent structure within the data to create meaningful positive and negative sample pairs. The detailed steps are as follows:

\begin{enumerate}
    \item \textbf{One-hot Encoding:} Convert all daily string representations into one-hot encoded vectors.
    \item \textbf{Clustering:} Apply K-means clustering to the one-hot encoded vectors to group similar daily patterns into clusters. 
    \item \textbf{Target Day Selection:} Choose a specific day as the target for comparison.
    \item \textbf{Similar Sample Selection:} For the target day, select a similar sample that meets all the following criteria:
        \begin{itemize}
            \item From the same participant
            \item Within a 30-day window of the target day
            \item Belongs to the same cluster as the target day
        \end{itemize}
    \item \textbf{Dissimilar Sample Selection:} Randomly select any other sample that does not meet the criteria for similar sample selection.
\end{enumerate}

We selected a 30-day interval for positive sample selection for two key reasons: first, k-means clustering of the encoded vectors yielded the best results with a 30-day window, as is shown in \ref{ablation}; second, many patients undergo regular physical checkups, such as urine tests, on a monthly basis, aligning well with this time frame.

\begin{table*}[htbp]
\centering
\caption{Silhouette scores under different models and parameter settings}
\label{ablation}
\begin{tabular}{l@{\hskip0.2in}r@{\hskip0.4in}r@{\hskip0.2in}r@{\hskip0.2in}r@{\hskip0.2in}r@{\hskip0.2in}}
\toprule
Model & Parameters & \multicolumn{4}{c}{Silhouette scores} \\
\cmidrule(r){3-6}
& & 4 & 5 & 6 & 7 \\
\midrule
MiniLM- & 7days & 0.459 & 0.451 & 0.431 & 0.413 \\
L12-v2  & 30days & \bfseries 0.554 & \bfseries 0.554 & \bfseries 0.554 &  \bfseries 0.542 \\
& 180days & 0.437 & 0.429 & 0.370 & 0.407 \\
\midrule
BAAI/bge- & 30days & 0.459 & 0.425 & 0.473 & 0.473 \\
small-v1.5\citep{xiao_c-pack_2023} & no tune & 0.173 & 0.165 & 0.181 & 0.170 \\
\bottomrule
\end{tabular}
\end{table*}

This ablation study is conducted to address potential concerns with our initial assumptions and to select the most suitable parameters that maximize the separation of different vector embeddings. By examining the k-means clustering results under various settings, we aim to identify the optimal configuration that yields the greatest distinction among the vector representations, mitigating potential issues arising from our assumptions.

In each epoch, we randomly selected 50,000 triplets from the dataset, using a batch size of 256. Sentence embeddings were evaluated using a triplet loss, where the Manhattan distance was calculated and optimized between the coding vectors of anchor samples and their corresponding positive and negative samples. Manhattan distance's suitability for sparse data, computational efficiency, and applicability in discrete systems make it a preferred choice for measuring similarity in our work. The loss function was optimized using the AdamW algorithm \citep{loshchilov_decoupled_2019} with a learning rate of $2 \times 10^{-5}$ and a weight decay of 0.01. A linear warm-up learning rate scheduler was applied with 10,000 warm-up steps. The training of large language model was carried out using NVIDIA A100 GPU, with each training epoch taking approximately 444.19 seconds.

\subsection{Location time series and distribution visualization for each cluster}
\begin{figure}[htbp]
    \centering
    \includegraphics[width=0.4\textwidth]{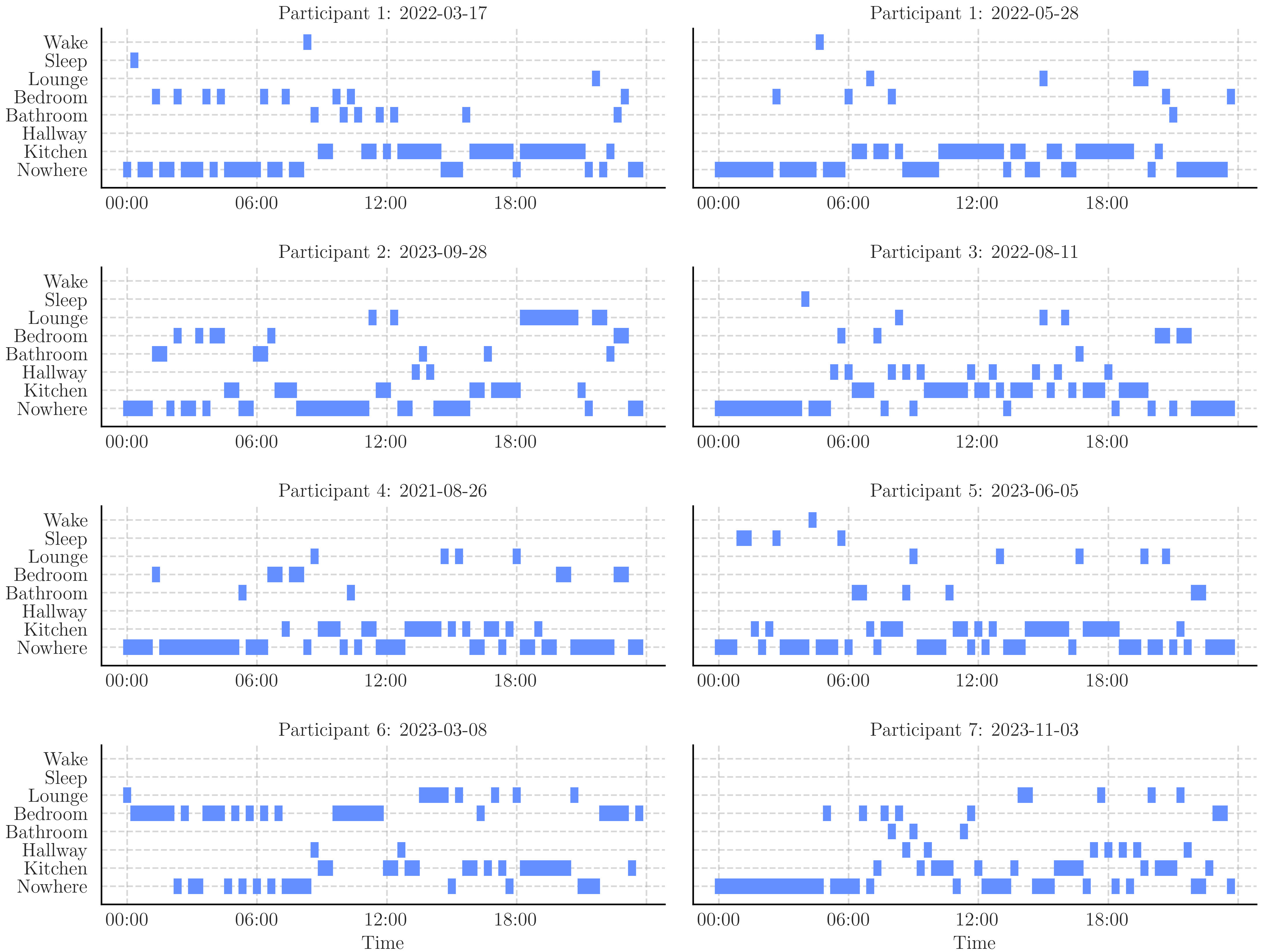}
    \caption{Location timeseries for articipants in cluster 1}
    \label{cluster1}
\end{figure}

\begin{figure}[htbp]
    \centering
    \includegraphics[width=0.4\textwidth]{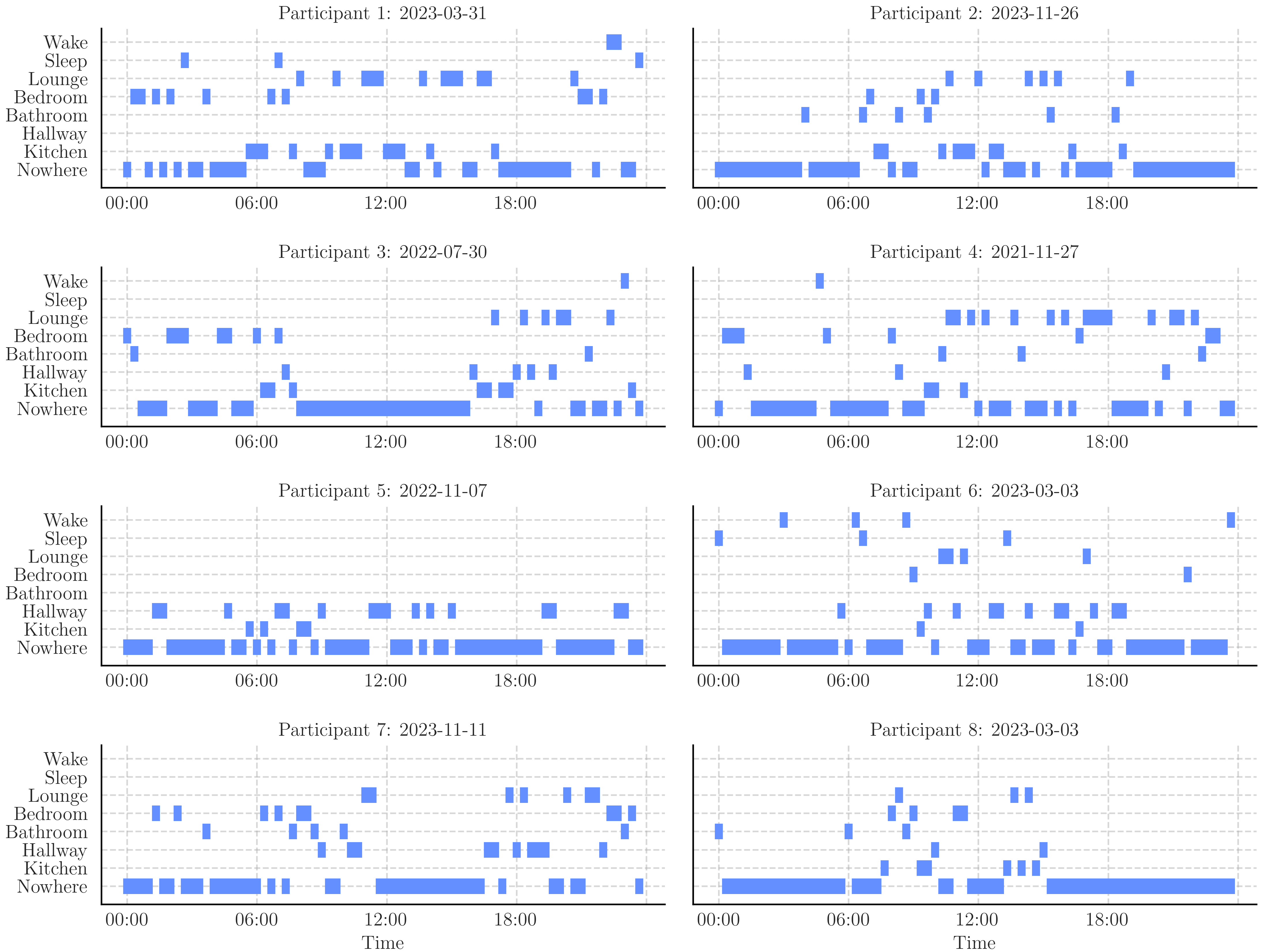}
    \caption{Location timeseries for participants in cluster 2}
    \label{cluster2}
\end{figure}

\begin{figure}[htbp]
    \centering
    \includegraphics[width=0.4\textwidth]{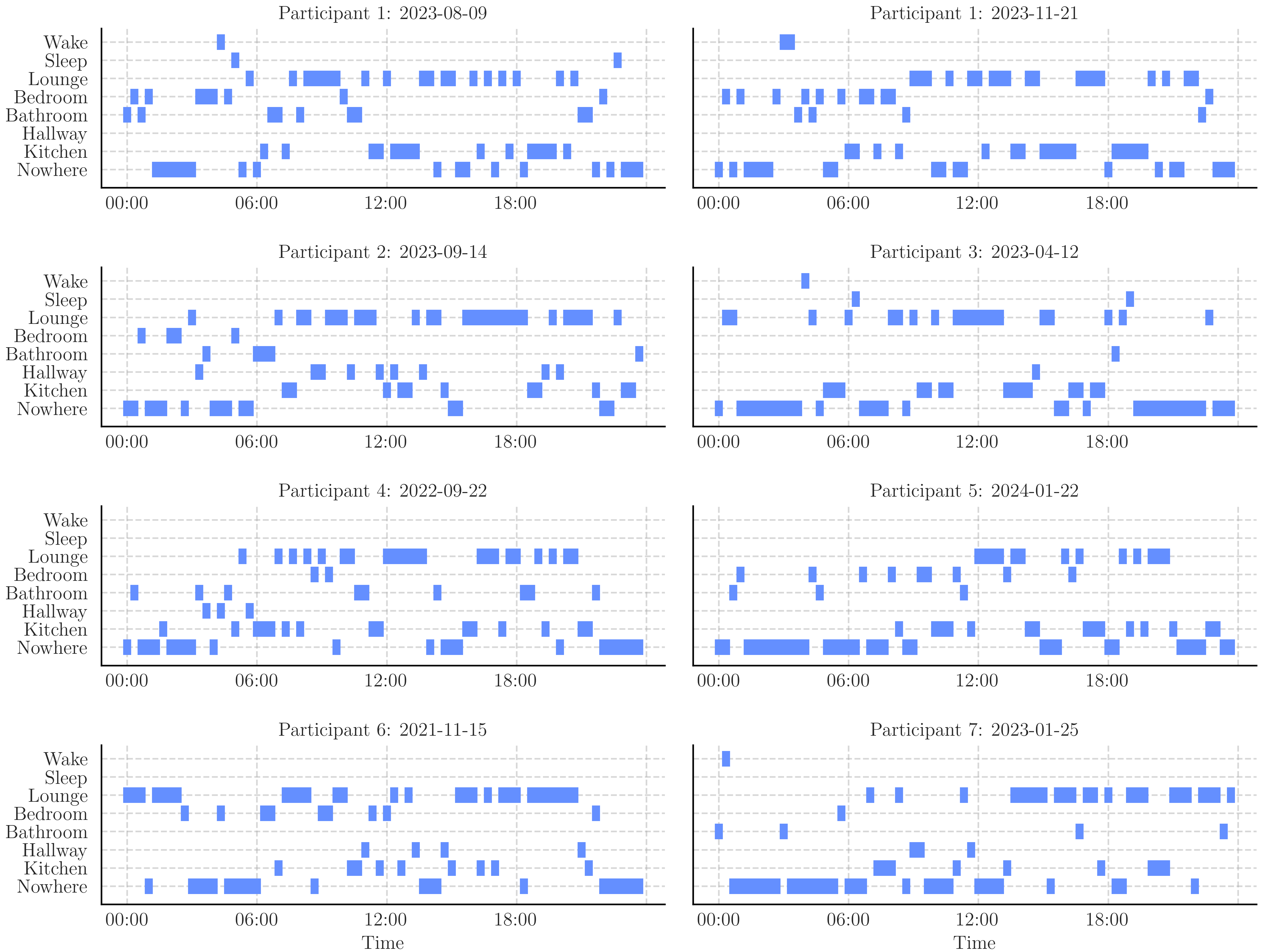}
    \caption{Location timeseries for participants in cluster 3}
    \label{cluster3}
\end{figure}

\begin{figure}[htbp]
    \centering
    \includegraphics[width=0.4\textwidth]{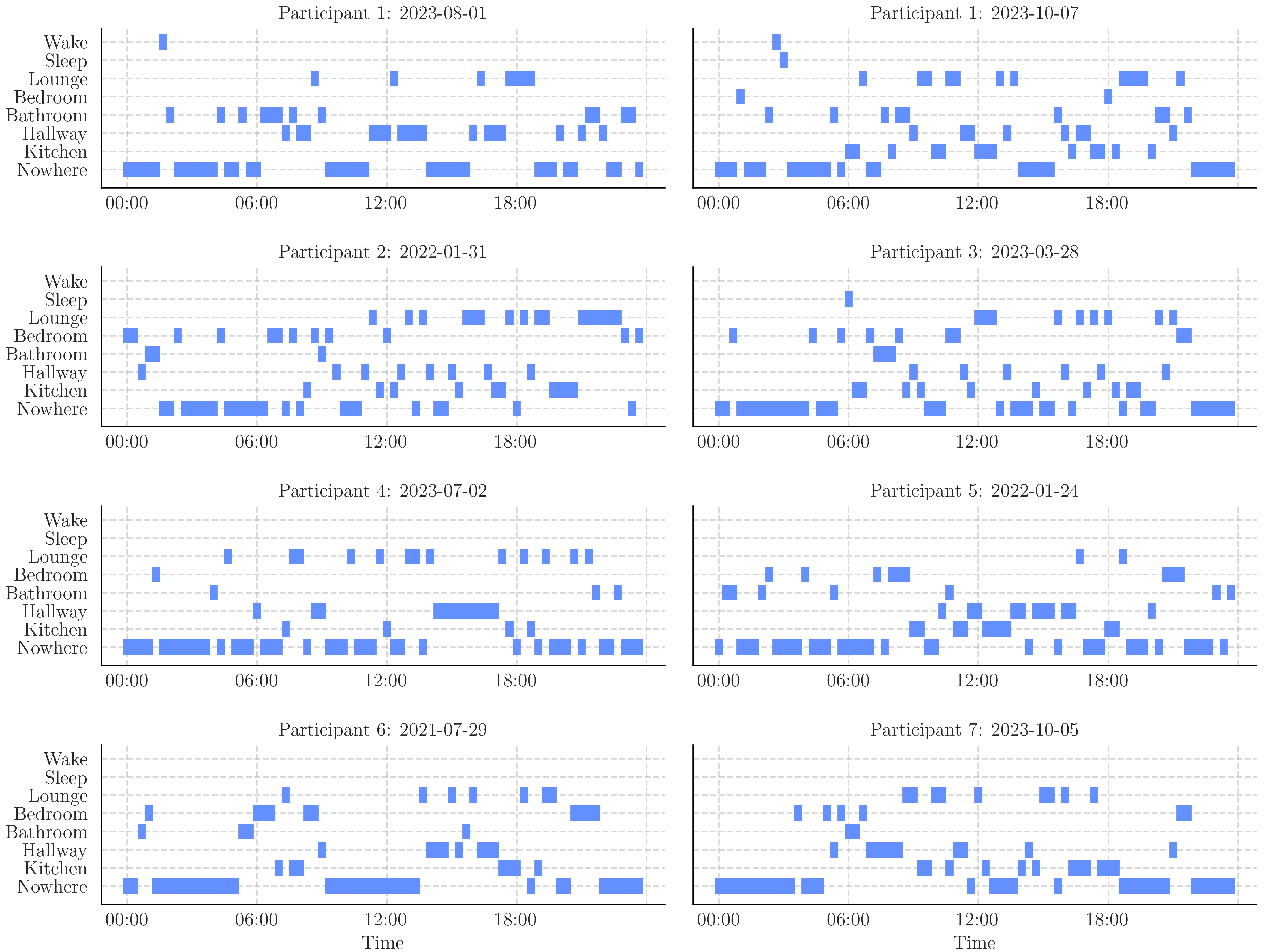}
    \caption{Location timeseries for participants in cluster 4}
    \label{cluster4}
\end{figure}

\begin{figure}[htbp]
    \centering
    \includegraphics[width=0.4\textwidth]{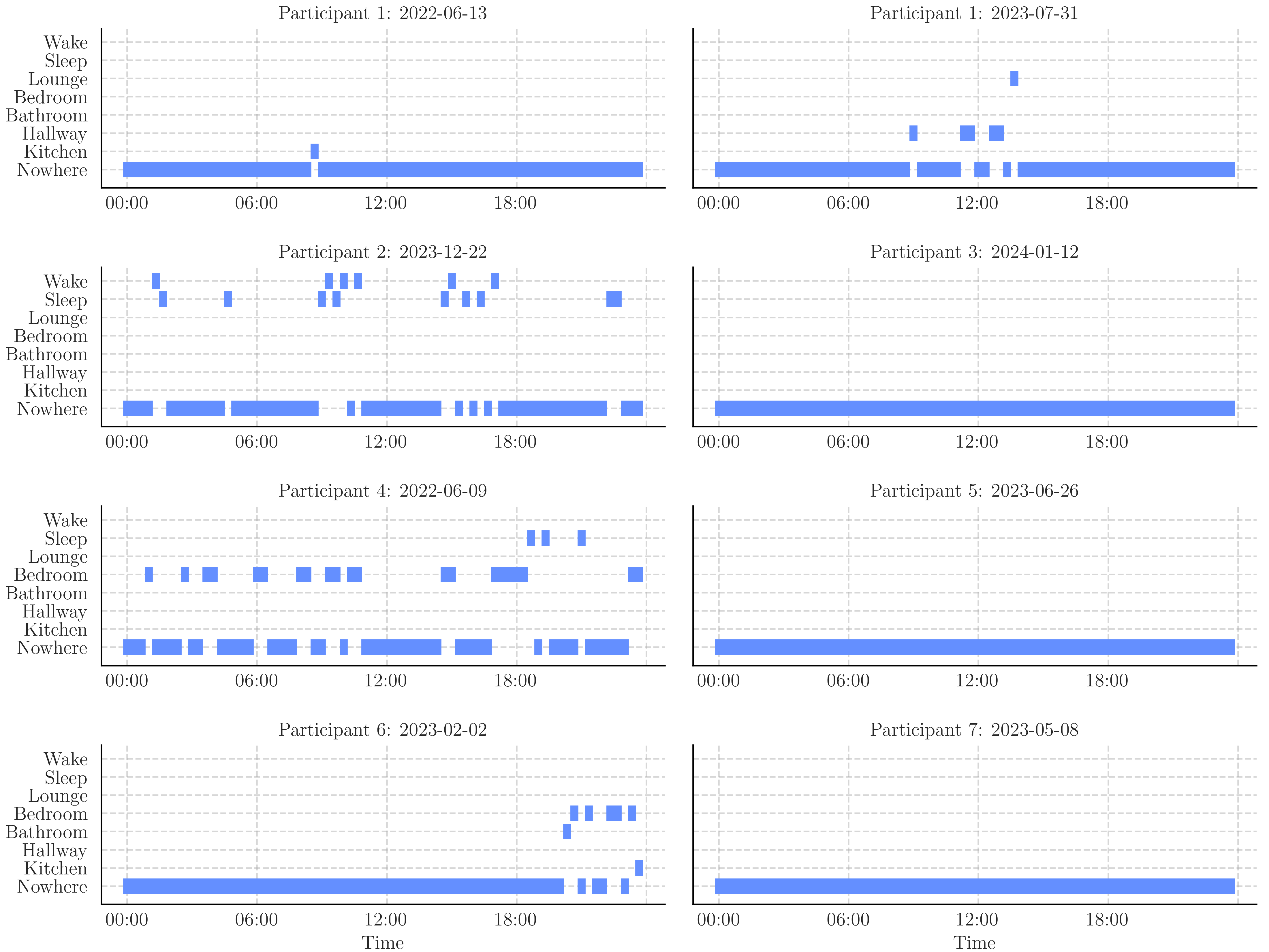}
    \caption{Location timeseries for participants in cluster 5}
    \label{cluster5}
\end{figure}

\begin{figure}[htbp]
    \centering
    \includegraphics[width=0.4\textwidth]{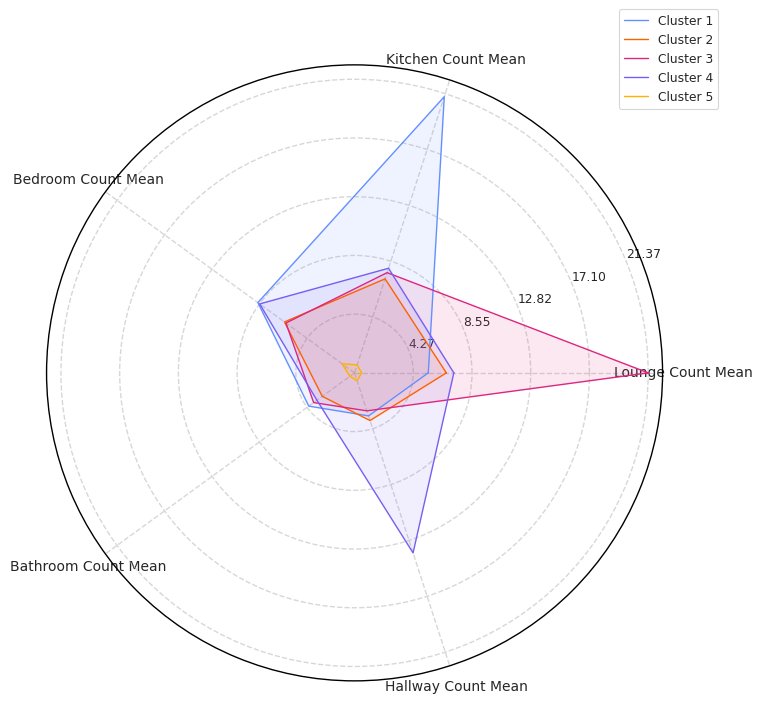}
    \caption{Mean of location count in five clusters}
    \label{clustermean}
\end{figure}

\begin{figure}[htbp]
    \centering
    \includegraphics[width=0.4\textwidth]{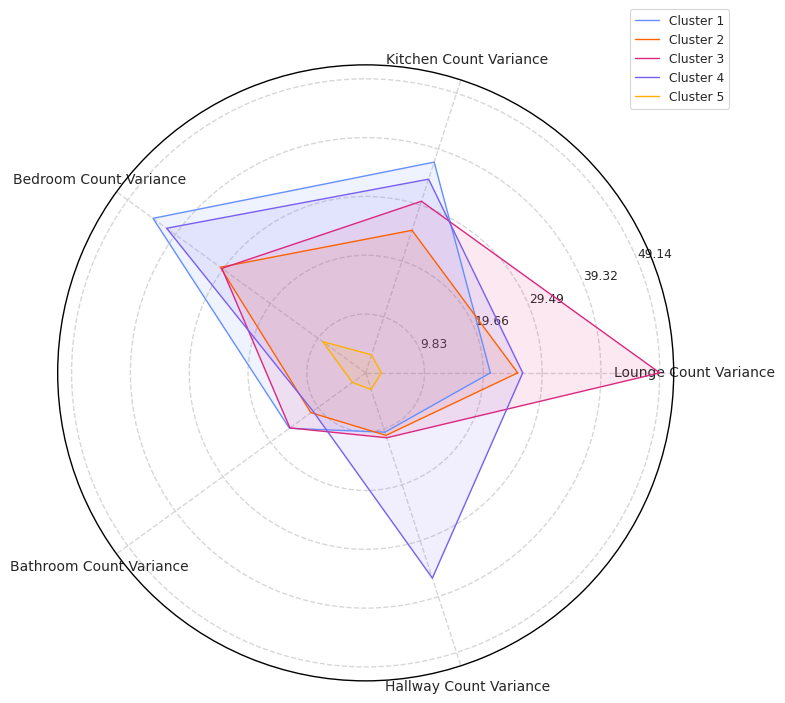}
    \caption{Variance of location count in five clusters}
    \label{clustervar}
\end{figure}

\begin{figure}[htbp]
    \centering
    \includegraphics[width=0.4\textwidth]{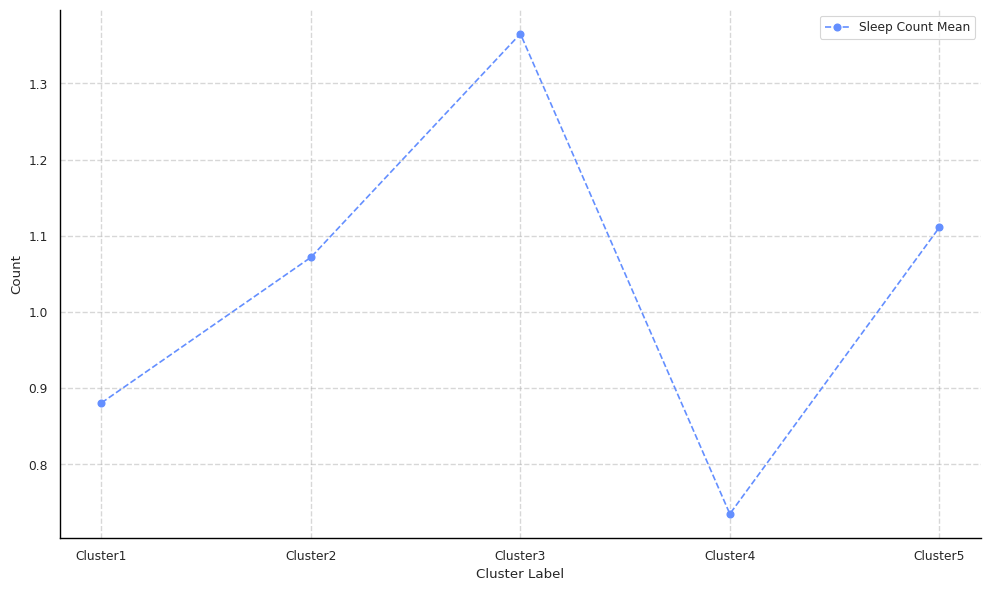}
    \caption{Mean of wake count in five clusters}
    \label{clusterwake}
\end{figure}

Following the extraction of location embeddings, we identified five clusters based on the highest silhouette score, achieved through k-means clustering. The resulting distribution presents the temporal clustering of each location for an individual participant, as is shown in Figure \ref{cluster1}, \ref{cluster2}, \ref{cluster3}, \ref{cluster4}, \ref{cluster5}, along with the cumulative distribution of each location across all participants in Figure \ref{clustermean}, \ref{clusterwake} and \ref{clustervar}.



\subsection{T-SNE plots for individual participants in test set}
\label{tsneplot2}

\begin{figure}[htbp]
    \centering
    \includegraphics[width=0.4\textwidth]{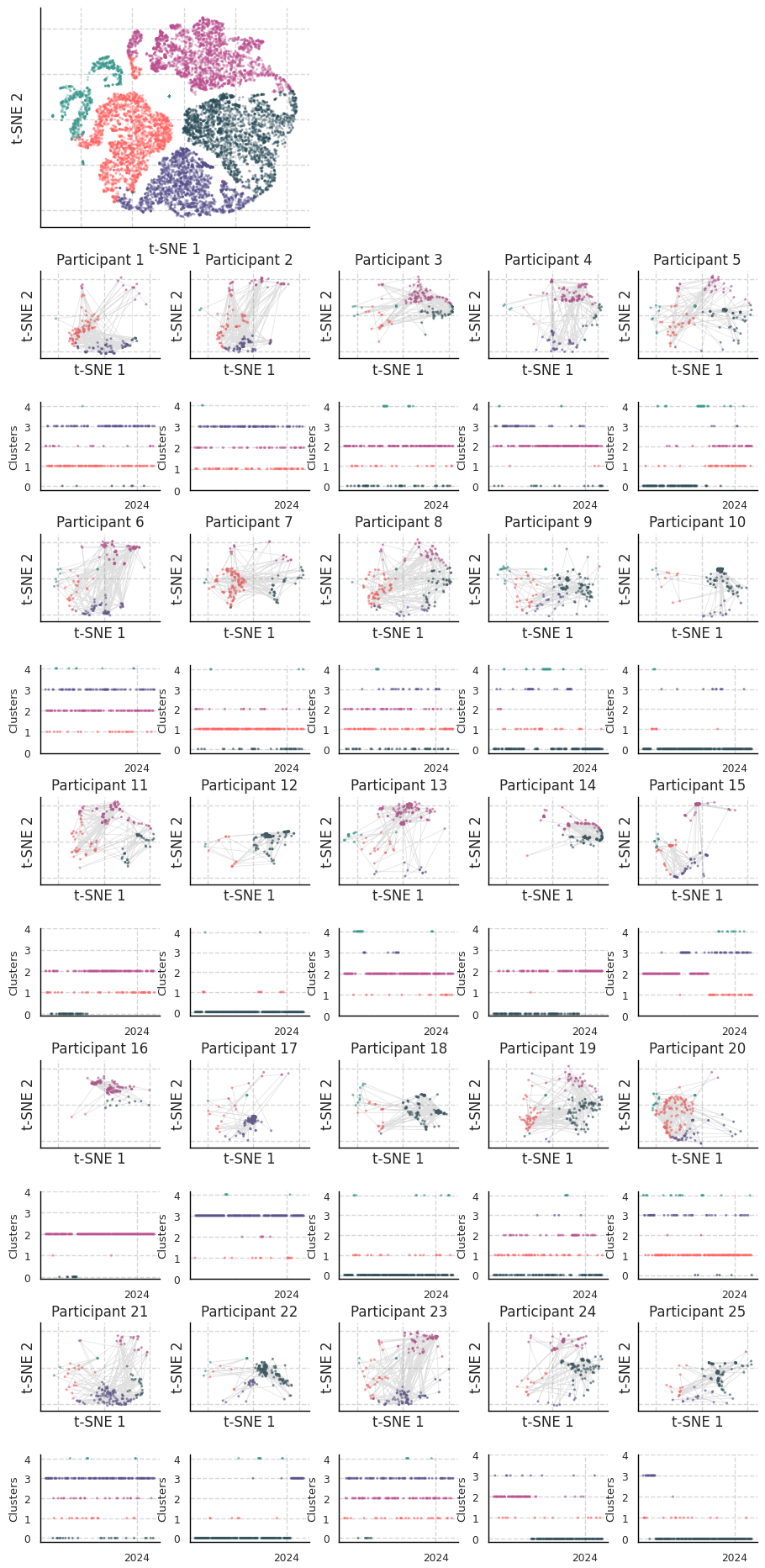}
    \caption{T-SNE for individuals in test set}
    \label{tsneplotpic2-1}
\end{figure}

\begin{figure}[htbp]
    \centering
    \includegraphics[width=0.4\textwidth]{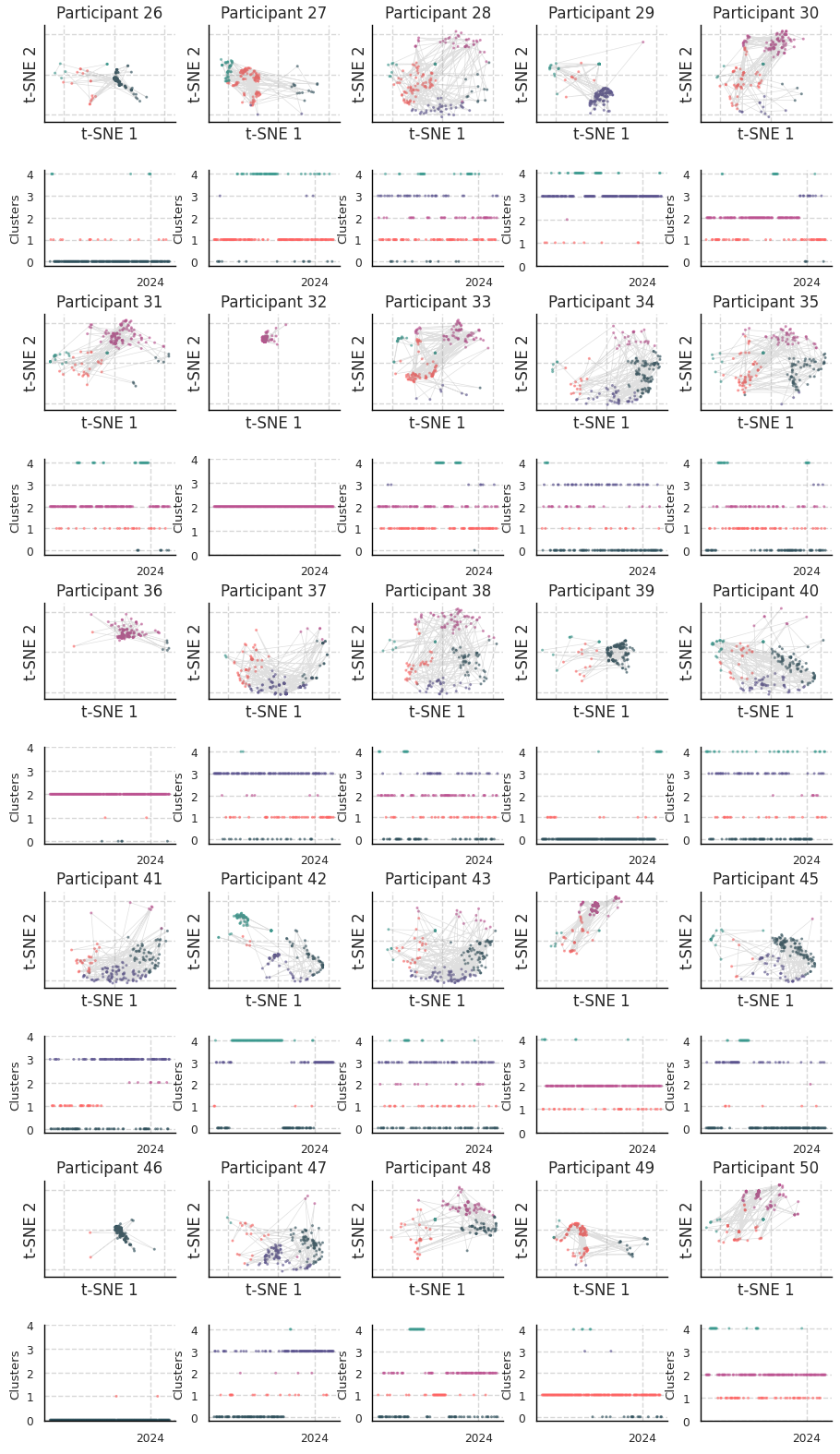}
    \caption{T-SNE for individuals in test set}
    \label{tsneplotpic2-2}
\end{figure}

Figure \ref{tsneplotpic2-1} and Figure \ref{tsneplotpic2-2} illustrate individual movement in embedded space, with their location and timespan. Every datapoint is collected after 2023-07-31 to 2024-01-31.

\subsection{PageRank Iteration for a Single Patient}
\label{pagerank}
\subsubsection{Model Definition}

We aim to compute the PageRank model fit and entropy value for a single patient based on their embeddings and cluster labels. The process involves defining a transition matrix based on distances between clusters, computing the PageRank scores.

\subsubsection{Transition Matrix Construction}

Given:
\begin{itemize}
    \item $X$: Patient embeddings (shape: $(n\_samples, 2)$).
    \item $y$: Patient cluster labels (shape: $(n\_samples,)$).
    \item $num\_clusters$: Number of clusters.
    \item $threshold$: Distance threshold for defining transitions.
\end{itemize}

The transition matrix $\mathbf{T}$ is computed as follows:

\begin{equation}
T_{ij} = \frac{\sum_{k \in C_i} \sum_{l \in C_j} \mathbb{1}_{\{d(k, l) \leq \text{threshold}\}}}{\sum_{l \in C_i} \sum_{m \in C_j} \mathbb{1}_{\{d(l, m) \leq \text{threshold}\}}}
\end{equation}

where:
\begin{itemize}
    \item $C_i$ and $C_j$ are the sets of samples in clusters $i$ and $j$, respectively.
    \item $d(k, l)$ denotes the distance between samples $k$ and $l$.
    \item $\mathbb{1}_{\{ \cdot \}}$ is an indicator function that equals 1 if the condition is true and 0 otherwise.
\end{itemize}

\subsubsection{PageRank Computation}

The PageRank vector $\mathbf{p}$ is computed iteratively using:

\begin{equation}
\mathbf{p}^{(t+1)} = \frac{1 - \alpha}{num\_clusters} + \alpha \mathbf{T}^\top \mathbf{p}^{(t)}
\end{equation}

where $\alpha$ is the damping factor (typically 0.85), and $\mathbf{T}^\top$ is the transpose of the transition matrix. The process continues until convergence:

\begin{equation}
\| \mathbf{p}^{(t+1)} - \mathbf{p}^{(t)} \|_1 < \text{tol}
\end{equation}

where $\text{tol}$ is a predefined tolerance for convergence.

\subsubsection{Algorithm Summary}
Algorithm~\ref{algorithm} explains the specific implementation steps of the PageRank we use. 

\begin{algorithm}
\caption{PageRank for a Single Patient}
\begin{algorithmic}
\STATE \textbf{Input:} $X$, $y$, $num\_clusters$, $threshold$, $\alpha$, $\text{max\_iter}$, $\text{tol}$
\STATE \textbf{Output:} $\mathbf{T}$, $\mathbf{p}$
\STATE Initialize transition matrix $\mathbf{T}$ with zeros
\FOR{each cluster $i$}
    \FOR{each cluster $j$}
        \STATE Compute distances between samples in clusters $i$ and $j$
        \STATE Update $\mathbf{T}_{ij}$ based on the distance threshold
    \ENDFOR
\ENDFOR
\STATE Normalize transition matrix $\mathbf{T}$
\STATE Initialize PageRank vector $\mathbf{p}$ uniformly
\FOR{iteration $t = 1$ to $\text{max\_iter}$}
    \STATE Compute new PageRank vector $\mathbf{p}^{(t+1)}$
    \IF{convergence condition met}
        \STATE Break
    \ENDIF
\ENDFOR
\STATE Compute PageRank matrix $\mathbf{P}_{\text{rank}}$
\end{algorithmic}
\label{algorithm}
\end{algorithm}

\subsubsection{Algorithm Visualization}
\label{pagerank_vis}

\begin{figure}[ht]
    \centering
    \subfigure{
        \includegraphics[width=0.2\textwidth]{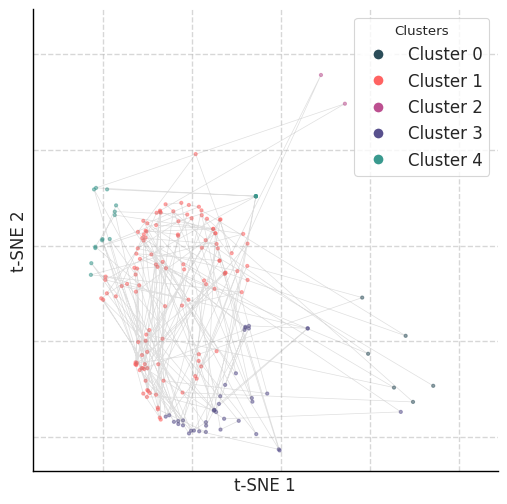}
        \label{fig:fig1}
    }
    \hfill
    \subfigure{
        \includegraphics[width=0.2\textwidth]{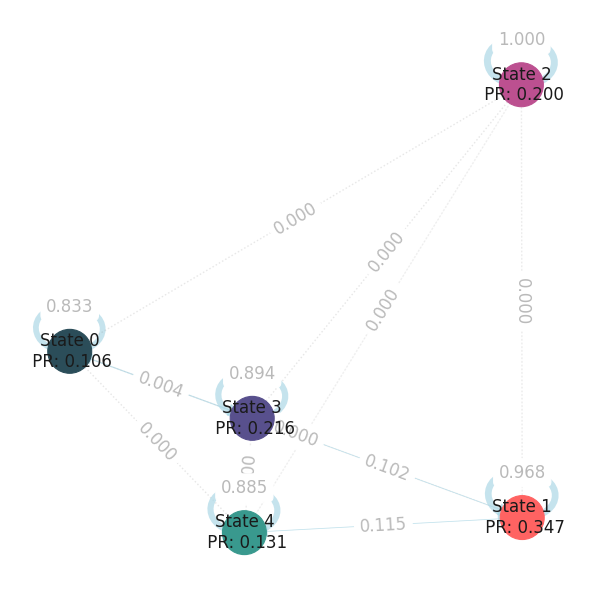}
        \label{fig:fig2}
    }
    \caption{Visualization of the generation of Pagerank value, left graph is single participant 2D t-SNE visualization, right graph is PageRank nodes value visualization}
    \label{fig:figures}
\end{figure}

Figure \ref{fig:figures} is a sample PageRank state generation process.

\subsection{Description of Feature Values Across Clusters Based on PageRank Vector Clustering}

\begin{figure}[htbp]
    \centering
    \includegraphics[width=0.4\textwidth]{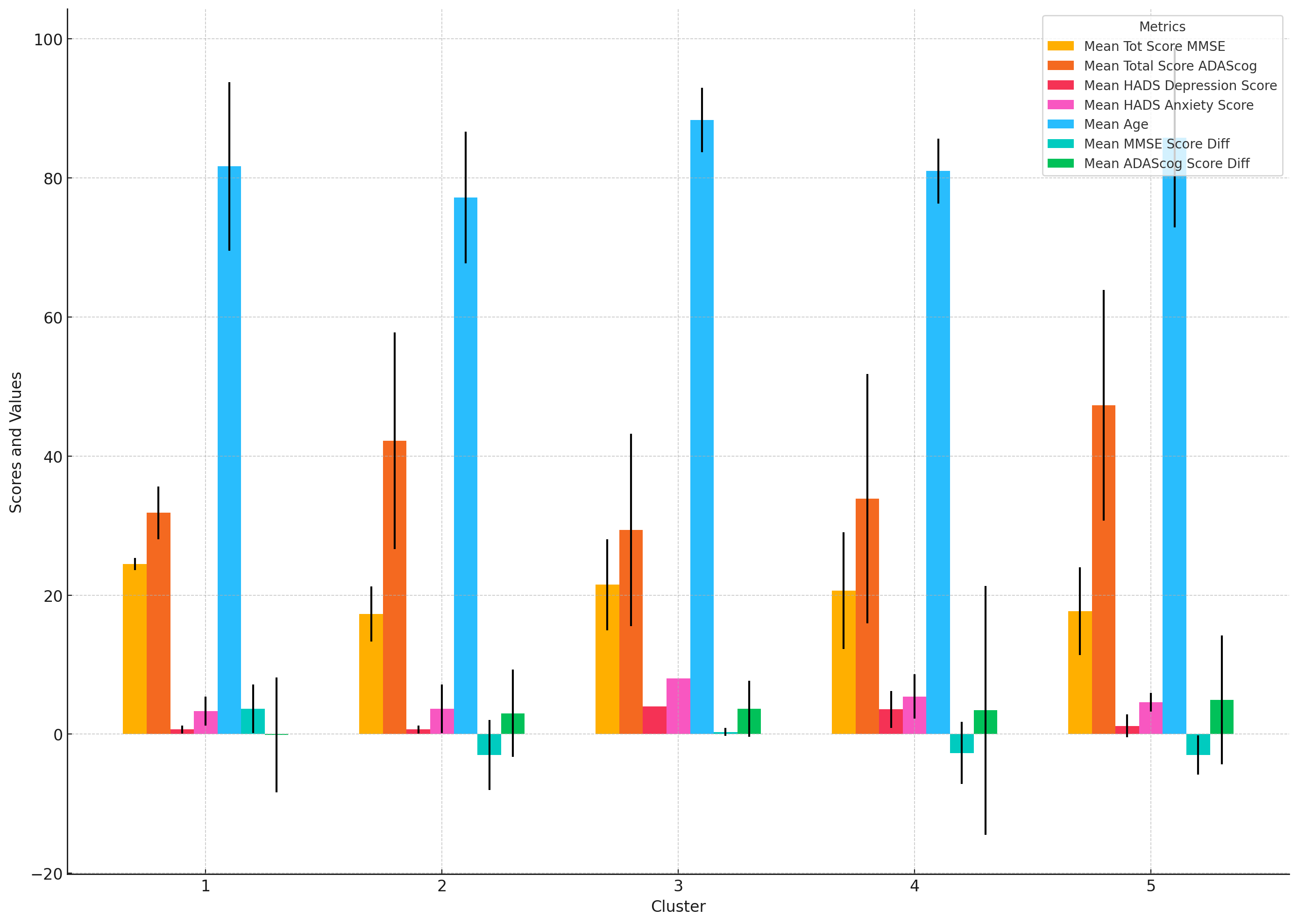}
    \caption{Feature values across clusters based on PageRank vector clustering}
    \label{clustering_pagerank}
\end{figure}

Figure \ref{clustering_pagerank} illustrates the mean values and distributions of various clinical and demographic metrics in five groups derived from the PageRank vector clustering process. It is a re-clustering process after the first clustering of language embedding and calculating PageRank vector for each patient. Right now each patient only have one vector with five elements in a certain time period, in this case 6 months. Each cluster is represented along the x-axis, with the corresponding mean values of different metrics displayed along the y-axis, accompanied by error bars to indicate variability.

The metrics included in this analysis are as follows:
\begin{itemize}
    \item \textbf{Mean Total Score MMSE}: This metric reflects cognitive function across clusters. Notably, Cluster 1 and Cluster 3 show higher MMSE scores, suggesting preserved or enhanced cognitive function relative to other clusters.
    \item \textbf{Mean Total Score ADAS-Cog}: The ADAS-Cog scores vary significantly across clusters. Clusters 2 and 5 have higher ADAS-Cog scores, indicating greater cognitive difficulties, while Clusters 3 and 4 demonstrate lower scores, possibly suggesting less impairment.
    \item \textbf{Mean HADS Depression Score}: Marked in magenta, this score captures levels of depression across clusters. Depression levels are relatively low in all clusters, with minimal variation, although clusters 3 and 4 show slight elevations.
    \item \textbf{Mean HADS Anxiety Score}: Represented in cyan, the anxiety levels are also uniformly low across clusters, with no notable inter-cluster variability except Cluster 3.
    \item \textbf{Mean Age}: The highest mean age appears in Cluster 3, followed closely by Cluster 5, indicating a possible association between age and specific behavioral patterns captured in these clusters.
    \item \textbf{Mean MMSE Score Difference}: This metric indicates changes in cognitive scores. Clusters 1 and 3 display higher positive differences, suggesting cognitive improvement, whereas Clusters 2, 4and 5 exhibit negligible or slightly negative values.
    \item \textbf{Mean ADAS-Cog Score Difference}: Shown in green, ADAS-Cog score changes vary across clusters, with Clusters 1 showing steadiness, whereas other clusters show slightly positive changes.
\end{itemize}

This clustering analysis demonstrates distinct patterns in clinical and demographic characteristics across the clusters, suggesting that the PageRank vector-based clustering approach effectively captures unique behavioral and cognitive profiles. These findings could facilitate more personalized assessments of cognitive health, with specific clusters potentially indicating varying degrees of cognitive resilience, vulnerability, and age-related changes.

\begin{figure}[htbp]
    \centering
    \includegraphics[width=0.4\textwidth]{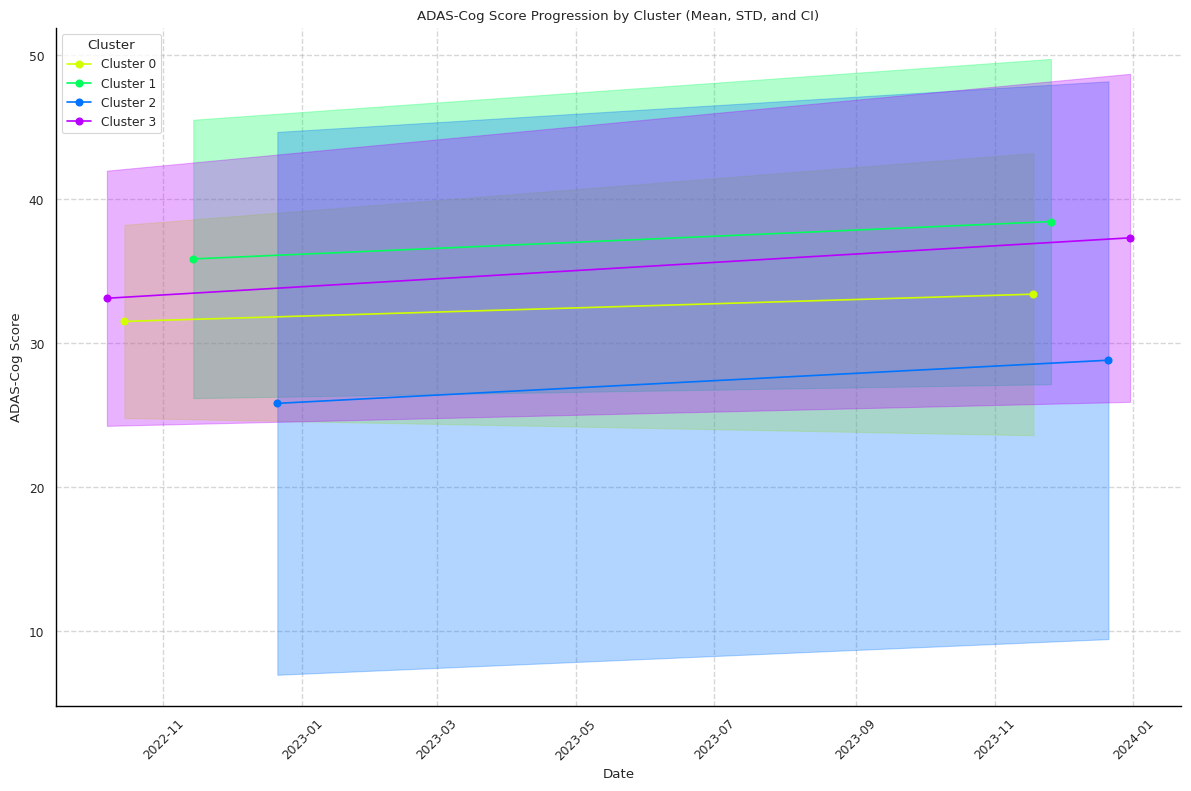}
    \caption{ADAS-Cog score progression by cluster (Mean, STD, and CI)}
    \label{clustering_pagerank_adas}
\end{figure}

\begin{figure}[htbp]
    \centering
    \includegraphics[width=0.4\textwidth]{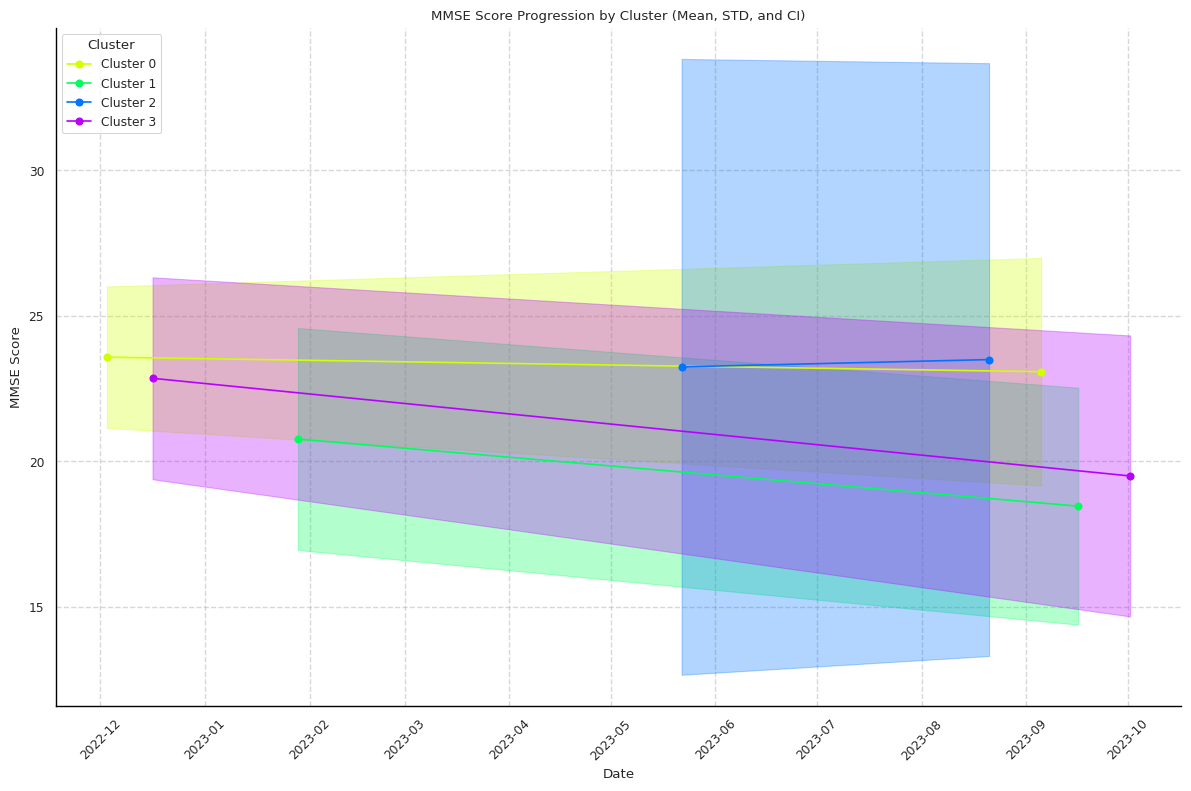}
    \caption{MMSE score progression by cluster (Mean, STD, and CI)}
    \label{clustering_pagerank_mmse}
\end{figure}

Figure \ref{clustering_pagerank_mmse}, \ref{clustering_pagerank_adas} illustrate the progression of cognitive scores—MMSE (Mini-Mental State Examination) and ADAS-Cog (Alzheimer's Disease Assessment Scale-Cognitive Subscale)—over time across different clusters, each representing a subset of participants grouped based on similar cognitive trajectory patterns.

Notably, for MMSE the scores generally decline over time, indicating a gradual cognitive decline. Clusters appear to vary in baseline MMSE scores and rate of decline, with some clusters demonstrating more stability and others a sharper reduction. For ADAS-Cog, clusters differ in terms of their baseline ADAS-Cog scores and the rate of increase over time. Some clusters show a relatively stable progression, while others exhibit a more rapid increase in scores, suggesting accelerated cognitive impairment. Together, these graphs highlight different cognitive trajectories between groups, showing PageRank vectors represents certain patterns of cognitive decline in different subgroups of participants.

\subsection{Analysis of PageRank-Derived Cognitive States}

\begin{figure}[htbp]
    \centering
    \includegraphics[width=0.4\textwidth]{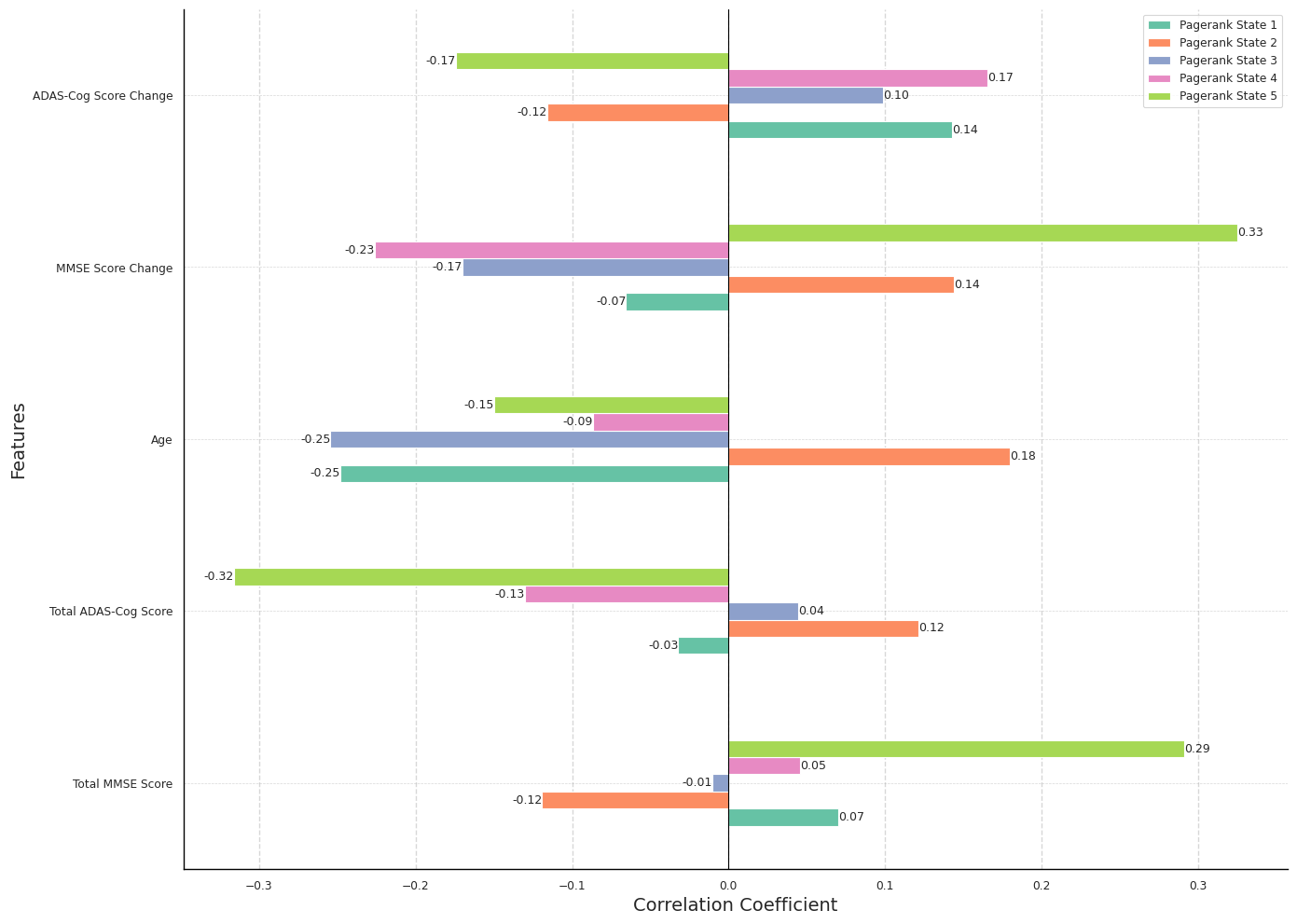}
    \caption{Correlation plot for PageRank states}
    \label{fig:CORRELATION}
\end{figure}

\begin{figure*}[htbp]
    \centering
    % 第一行子图
    \begin{subfigure}
        \centering
        \includegraphics[width=0.3\textwidth]{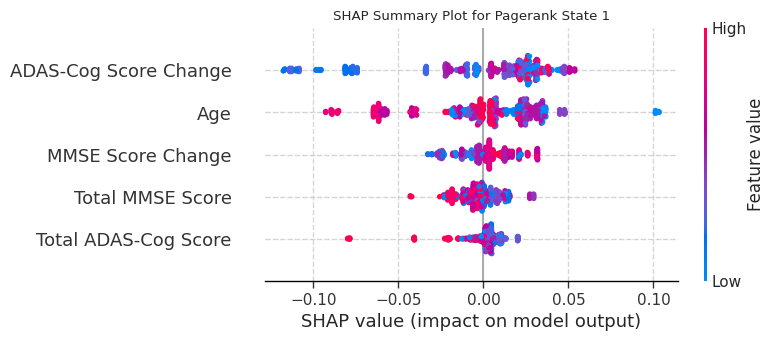}
        \label{fig:shap_state1}
    \end{subfigure}
    \hfill
    \begin{subfigure}
        \centering
        \includegraphics[width=0.3\textwidth]{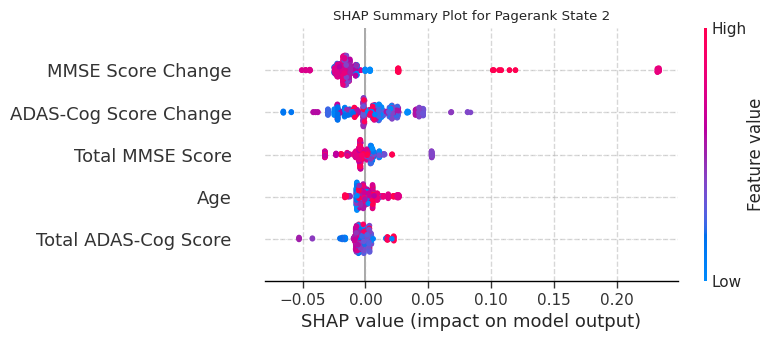}
        \label{fig:shap_state2}
    \end{subfigure}
    \hfill
    \begin{subfigure}
        \centering
        \includegraphics[width=0.3\textwidth]{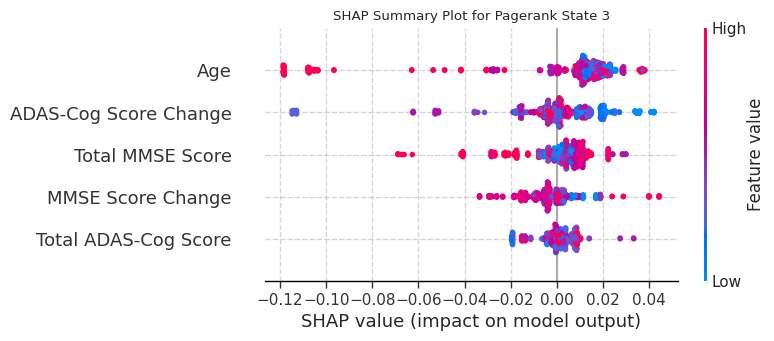}
        \label{fig:shap_state3}
    \end{subfigure}

    % 第二行子图
    \begin{subfigure}
        \centering
        \includegraphics[width=0.3\textwidth]{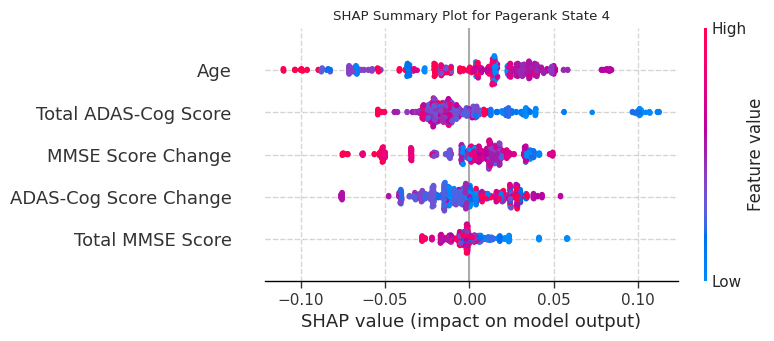}
        \label{fig:shap_state4}
    \end{subfigure}
    \hfill
    \begin{subfigure}
        \centering
        \includegraphics[width=0.3\textwidth]{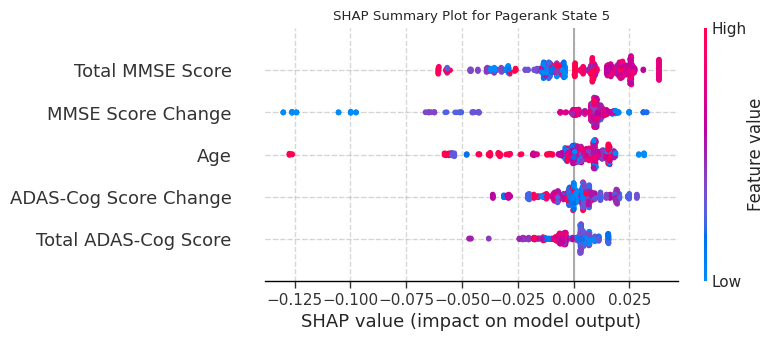}
        \label{fig:shap_state5}
    \end{subfigure}

    \caption{SHAP summary plots for PageRank states 1 to 5}
    \label{fig:shap_summary_plots}
\end{figure*}

Our analysis of five PageRank-derived states, representing distinct behavioral patterns, reveals:

\begin{enumerate}
    \item \textbf{Feature Importance Variation:} The significance of cognitive and functional measures varies considerably across states in Figure \ref{fig:CORRELATION} and \ref{fig:shap_summary_plots}, indicating distinct behavioral characteristics.
    
    \item \textbf{ADAS-Cog Score Change:} 
    \begin{itemize}
        \item State 1,3 and 4 shows positive impact, suggesting cognitive decline.
        \item State 5 and 2 exhibits strong negative correlation (-0.17 and -0.12), potentially indicating cognitive stability or improvement.
    \end{itemize}
    
    \item \textbf{MMSE Score Change:}
    \begin{itemize}
        \item Highest variability in State 2, implying significant changes in global cognitive function.
        \item Strong positive correlation (0.33) in State 5, further supporting cognitive improvement hypothesis.
    \end{itemize}
    
    \item \textbf{Age Effects:}
    \begin{itemize}
        \item Positive impact in State 2, suggesting prevalence in older participants.
        \item Negative correlations in States 1 and 3 (-0.25 and -0.25), indicating patterns more characteristic of younger participants.
    \end{itemize}
    
    \item \textbf{Total Cognitive Scores:}
    \begin{itemize}
        \item Complex relationship between Total ADAS-Cog Score and State 2.
        \item Strong positive correlation (0.29) of Total MMSE Score in State 5, supporting better cognitive function.
    \end{itemize}

    \item \textbf{Looking back to location counts:}
    \begin{itemize}
        \item Shown in Figure \ref{clustermean} and \ref{clustervar}, analysis of the mean and variance plots for location counts reveals distinct patterns across cognitive states. In State 1, the frequency of kitchen visits is markedly elevated, significantly exceeding the mean observed in other states. This high kitchen usage may suggests possible cognitive decline and disruption in daily routines, a hypothesis further supported by changes in MMSE scores. For State 2, the frequency of each location remains relatively balanced, indicating a stable cognitive state, consistent with our assessments. In State 3, the lounge shows a high usage frequency with large variance, implying that the participant maintains considerable mobility and regularly relaxes in the living room, aligning with the relatively younger age of this individual based on our calculations. State 4 is characterized by an unusually high frequency of hallway usage, which may indicate cognitive impairment, as corroborated by the MMSE score trajectory. This elevated hallway frequency suggests the participant may have difficulty recalling the room layout, resulting in aimless movement within the home. Finally, in State 5, location frequencies are low and approach zero, potentially due to equipment malfunction or the participant’s absence from the home. This pattern may also indicate that the participant is frequently outside for more than one day, reflecting a relatively healthy and independent cognitive state if they can engage in extended outdoor activities.
    \end{itemize}
\end{enumerate}

\subsubsection{State-Specific Observations}
\begin{itemize}
    \item \textbf{State 1: "Transition Period"}
    \begin{itemize}
        \item Possibly represents an early stage of mild cognitive decline.
        \item Younger population with subtle cognitive changes, potentially a mix of normal aging and early pathological changes.
    \end{itemize}

    \item \textbf{State 2: "Stable Elderly Period"}
    \begin{itemize}
        \item Likely represents a healthy elderly population or patients with stable chronic conditions.
        \item Older age but relatively stable cognitive function; ADAS-Cog score changes closely related to state.
    \end{itemize}

    \item \textbf{State 3: "High-Risk Youth Period"}
    \begin{itemize}
        \item May represent a high-risk group for early-onset cognitive disorders.
        \item Cognitive fluctuations in younger population, especially in MMSE score changes.
    \end{itemize}

    \item \textbf{State 4: "Rapid Progression Period"}
    \begin{itemize}
        \item Possibly a stage of dramatic cognitive decline.
        \item Elderly group with significant changes in cognitive test scores; unstable state.
    \end{itemize}

    \item \textbf{State 5: "Stable Youth Period"}
    \begin{itemize}
        \item Cognitive stability or improvement, particularly in younger participants.
    \end{itemize}
\end{itemize}

These findings highlight the heterogeneity of cognitive decline patterns and may inform more personalized approaches to cognitive assessment and intervention in clinical practice.

This analysis highlights the potential of PageRank-derived states as markers of underlying cognitive behavior patterns, paving the way for individualized intervention strategies based on behavioral state identification.

\subsection{Clinical Significance of Deep Vector States}\label{secquan}

Through detailed analysis of the clinical relevance of deep vector states, their clustering, and patient behavior similarity, we have discovered that these deep but compact vectors encapsulate substantial information about patient behavior patterns. Each state represents a distinct cognitive or behavioral profile, with the intensity of each state reflecting specific cognitive patterns. Our objective now is to explore the potential clinical applications of these states by using them as features in machine learning models to predict MMSE and ADAS-Cog. These models will be trained by combining the vector states with baseline models and additional patient characteristics to assess their predictive performance.

The features included in the analysis are categorized as follows:

\begin{itemize}
    \item \textbf{Characteristics: Clinical Assessment Features}: This includes traditional clinical scores such as the \textit{HADS - Depression Score}, \textit{HADS - Anxiety Score}, clinical cognitive scores such as \textit{MMSE Score, ADAS-Cog Score(while predicting changes of these scores)}  and \textit{Age}.
    \item \textbf{Characteristics: Demographic and Diagnostic Features}: Non-numeric features like \textit{Gender}, \textit{Living Status (Alone or with others)}, and \textit{Clinical Diagnosis(dementia type)}.
    \item \textbf{Latent State Features}: The five latent states derived from deep vector representations (\textit{state1}, \textit{state2}, \textit{state3}, \textit{state4}, \textit{state5}) which capture essential cognitive and behavioral dynamics.
    \item \textbf{Baseline Features}: Aggregated time-series features, including daily mean and variance of movement across various household locations such as \textit{Wake Count}, \textit{Lounge Count}, \textit{Kitchen Count}, \textit{Bedroom Count}, \textit{Bathroom Count}, \textit{Hallway Count}, and \textit{Sleep Count}. Each location contributes two variables: the mean and variance of visit counts.
    \item \textbf{Frequency-Based Baseline:} This approach calculates the occurrence frequency of various activity locations for each participant, normalizing the count data each day to represent the percentage of time spent in different locations. For each location (\textit{e.g., Bedroom, Kitchen, Lounge}), statistical features such as mean and variance are computed for these normalized frequencies. These features serve as inputs to predictive models.

    \item \textbf{Random Word Baseline:} In this method, each location or activity state is mapped to a random numeric value. These random numeric embeddings replace meaningful semantic information and are used as features for model training. This baseline provides a reference point to evaluate the significance of structured embeddings or semantic features derived from location sequences.
    \item \textbf{Outcome Features}: The target variables for prediction, which are changes in clinical cognitive scores such as \textit{MMSE Score, \(\Delta\)MMSE, ADAS-Cog Score} and  \textit{\(\Delta\)ADAS-Cog}.
\end{itemize}

The baseline model was constructed to align closely with the dimensionality of the latent state features, otherwise, the scale of movement data baseline features for a single participant would be excessively large (7,776,000) and impossible to use. This baseline feature set aggregates time-series data by taking the mean and variance of daily activity at different locations within the home during the testing period. When predicting cognitive changes over time, the model could allows the inclusion of current values to enhance the prediction.Note that the length of the time-series input is six month in this case.

\subsubsection{Model Parameters}

To analyze the performance of the models, we employed three algorithms: XGBoost, LightGBM, and Support Vector Machine (SVM). The specific configurations of each model are as follows:

\begin{itemize}
    \item \textbf{XGBoost}: Key parameters include:
    \begin{itemize}
        \item \textit{n\_estimators}: 100, specifying the number of boosting rounds.
        \item \textit{learning\_rate}: 0.1, controlling the step size shrinkage to prevent overfitting.
        \item \textit{max\_depth}: 6, setting the maximum depth of each tree.
        \item \textit{random\_state}: 0, ensuring reproducibility.
    \end{itemize}
    
    \item \textbf{LightGBM}: Configured with the following parameters:
    \begin{itemize}
        \item \textit{n\_estimators}: 100, specifying the number of boosting iterations.
        \item \textit{learning\_rate}: 0.1, controlling the contribution of each tree.
        \item \textit{max\_depth}: -1, allowing the model to grow trees without a maximum depth constraint.
        \item \textit{random\_state}: 0, ensuring consistent results.
    \end{itemize}
    
    \item \textbf{Support Vector Machine (SVM)}: Configured as follows:
    \begin{itemize}
        \item \textit{kernel}: 'rbf', employing a radial basis function kernel for nonlinear decision boundaries.
        \item \textit{C}: 1.0, setting the regularization parameter.
        \item \textit{epsilon}: 0.1, defining the tolerance for error in the prediction.
    \end{itemize}
\end{itemize}

The machine learning models were trained using data collected from 50 participants. A variety of feature sets were explored, including baseline activity features, clinical characteristics, and derived latent states, as well as combinations thereof. For preprocessing, numerical features were standardized, while categorical features were processed with the most frequent strategy for imputation followed by one-hot encoding. 

The modeling approach included XGBoost, LightGBM, and SVM, selected for their ability to capture both linear and nonlinear relationships. Leave-One-Out Cross-Validation (LOOCV) was employed to assess the models' performance due to the small sample size, ensuring robustness and generalizability of the results. During each fold, the data was split into a single test sample and the remaining samples for training. Model evaluation focused on two primary metrics: Mean Absolute Error (MAE) and Root Mean Squared Error (RMSE), calculated separately for MMSE and ADAS-Cog score predictions.

\subsection{LSTM Model for Multivariate Time Series Prediction}
\textbf{Input:} The input to the LSTM model consists of weekly aggregated time series features derived from patient activity data. Each feature represents the count of occurrences for various activity locations (\textit{e.g., Bedroom, Kitchen, Lounge}) within a week. These time series data are standardized using \texttt{StandardScaler} to normalize the feature values across different scales.

\textbf{Model Architecture:}
\begin{itemize}
    \item \textbf{LSTM Layer:} A single LSTM layer with 64 units is used to capture temporal dependencies in the input data. This layer outputs a fixed-length feature vector summarizing the temporal patterns.
    \item \textbf{Dropout Layer:} A dropout layer with a rate of 0.2 is added to prevent overfitting by randomly setting a fraction of input units to zero during training.
    \item \textbf{Dense Layers:} Two fully connected dense layers are used:
    \begin{itemize}
        \item A hidden dense layer with 64 units and ReLU activation.
        \item An output dense layer with a size matching the number of target variables (\textit{e.g., MMSE, ADAS-COG scores and their differences}).
    \end{itemize}
\end{itemize}

\textbf{Training Strategy:}
\begin{itemize}
    \item \textbf{Loss Function:} Mean Squared Error (MSE) is used as the loss function for training.
    \item \textbf{Optimizer:} The Adam optimizer is employed for efficient gradient descent.
    \item \textbf{Validation:} 10-fold cross-validation is applied to evaluate the model's performance. In each fold, the data is split into training and testing sets.
    \item \textbf{Hyperparameters:} The model is trained for 50 epochs with a batch size of 32 in each fold.
\end{itemize}

\begin{table*}[htbp]
\caption{Performance metrics for predicting changes of ADAS-Cog and MMSE with 95\% confidence intervals (CI). The values in parentheses represent the lower and upper bounds of the 95\% CI. The metrics represent the predicted annual change (\(\Delta\)) for MMSE and ADAS-Cog.}
\resizebox{\textwidth}{!}{
\begin{tabular}{llllll}
\toprule
 & \textbf{Metric} & \textbf{MAE\textsubscript{\(\Delta\)ADASCOG}} & \textbf{MAE\textsubscript{\(\Delta\)MMSE}} & \textbf{RMSE\textsubscript{\(\Delta\)ADASCOG}} & \textbf{RMSE\textsubscript{\(\Delta\)MMSE}} \\
Model & Feature Set &  &  &  &  \\
\midrule
\multirow[t]{13}{*}{LightGBM} & All Features & 5.60 (4.24, 7.14) & 2.37 (1.78, 3.04) & 5.64 (4.20, 7.05) & 2.34 (1.73, 2.97) \\
 & Baseline & 6.67 (4.84, 8.70) & 2.57 (1.83, 3.37) & 6.76 (5.05, 8.72) & 2.57 (1.83, 3.37) \\
 & Characteristics & \textbf{4.96 (3.59, 6.50)} & 2.26 (1.72, 2.90) & \textbf{4.99 (3.67, 6.58)} & 2.24 (1.67, 2.89) \\
 & Proportion Baseline & 6.12 (4.61, 7.89) & 2.65 (2.04, 3.38) & 6.14 (4.57, 7.87) & 2.64 (2.04, 3.31) \\
 & Random Word & 5.69 (4.27, 7.47) & 2.75 (2.10, 3.47) & 5.69 (4.26, 7.36) & 2.74 (2.13, 3.46) \\
 & State & 6.00 (4.41, 7.75) & 2.70 (2.04, 3.37) & 6.06 (4.56, 7.83) & 2.69 (2.06, 3.42) \\
\cline{1-6}
\multirow[t]{13}{*}{SVM} & All Features & 6.29 (4.77, 8.09) & 3.60 (2.98, 4.29) & 6.27 (4.76, 8.21) & 3.59 (2.92, 4.35) \\
 & Baseline & 6.26 (4.68, 8.03) & 3.58 (2.97, 4.26) & 6.27 (4.75, 8.04) & 3.62 (2.99, 4.37) \\
 & Characteristics & 6.19 (4.62, 7.96) & 3.59 (2.96, 4.29) & 6.28 (4.77, 8.12) & 3.61 (3.04, 4.30) \\
 & Proportion Baseline & 6.26 (4.77, 8.05) & 3.61 (2.96, 4.33) & 6.28 (4.74, 8.02) & 3.60 (2.98, 4.34) \\
 & Random Word & 6.26 (4.71, 7.76) & 3.58 (3.01, 4.31) & 6.25 (4.64, 7.96) & 3.60 (2.96, 4.33) \\
 & State & 6.23 (4.67, 8.01) & 3.60 (2.98, 4.32) & 6.25 (4.63, 7.94) & 3.61 (2.97, 4.34) \\
\cline{1-6}
\multirow[t]{13}{*}{XGBoost} & All Features & 6.22 (4.85, 7.91) & 2.18 (1.54, 2.92) & 6.24 (4.67, 7.88) & 2.16 (1.60, 2.87) \\
 & Baseline & 6.13 (4.43, 8.02) & 2.91 (2.10, 3.79) & 6.11 (4.37, 7.82) & 2.92 (2.07, 3.78) \\
 & Characteristics & 5.60 (3.97, 7.49) & \textbf{2.05 (1.42, 2.77)} & 5.61 (4.08, 7.36) & \textbf{2.08 (1.44, 2.82)} \\
 & Proportion Baseline & 6.98 (5.38, 8.77) & 3.44 (2.72, 4.26) & 7.02 (5.25, 8.79) & 3.47 (2.71, 4.26) \\
 & Random Word & 7.16 (5.41, 9.21) & 2.82 (1.97, 3.69) & 7.18 (5.46, 9.29) & 2.84 (1.97, 3.72) \\
 & State & 7.30 (5.60, 9.22) & 3.33 (2.57, 4.14) & 7.31 (5.55, 9.35) & 3.33 (2.58, 4.13) \\
\cline{1-6}
\bottomrule
\end{tabular}}
\label{change_current}
\end{table*}

The performance metrics in Table \ref{change_current} summarize the predictive accuracy of various models for forecasting annual changes in ADAS-Cog and MMSE scores (\(\Delta\)ADASCOG and \(\Delta\)MMSE, respectively).

For the \textbf{LightGBM} model, the lowest prediction errors for ADAS-Cog were achieved using the feature set of \textit{Random Word + Characteristics}, with an MAE\textsubscript{\(\Delta\)ADASCOG} of \textbf{4.95} (95\% CI: 3.58, 6.42). For MMSE, the lowest prediction errors were observed with the \textit{State + Characteristics} feature set, resulting in an MAE\textsubscript{\(\Delta\)MMSE} of \textbf{2.24} (95\% CI: 1.67, 2.89). These findings suggest that focusing on well-selected clinical features and random word embeddings or combining state and clinical characteristics can improve predictive accuracy for both metrics.

For the \textbf{SVM} model, the performance was generally less favorable compared to LightGBM and XGBoost. The best-performing feature set was \textit{All Features}, with an MAE\textsubscript{\(\Delta\)ADASCOG} of 6.29 (95\% CI: 4.77, 8.09) and an MAE\textsubscript{\(\Delta\)MMSE} of 3.60 (95\% CI: 2.98, 4.29). Despite the inclusion of all feature types, SVM exhibited higher error margins, particularly for MMSE predictions.

For the \textbf{XGBoost} model, the combination of \textit{State + Random Word + Characteristics} yielded the most accurate predictions for MMSE, achieving an MAE\textsubscript{\(\Delta\)MMSE} of \textbf{2.02} (95\% CI: 1.44, 2.76). Similarly, the feature set of \textit{Characteristics} resulted in the best performance for ADAS-Cog, with an MAE\textsubscript{\(\Delta\)ADASCOG} of \textbf{5.41} (95\% CI: 4.07, 7.11). These results highlight the effectiveness of leveraging state and clinical features alongside embeddings for accurate prediction of cognitive changes.

As seen in Table \ref{change_current}, the combination of clinical characteristics and specific feature augmentations, such as random words or state embeddings, consistently improved predictive accuracy, particularly for MMSE. However, baseline models and standalone state features generally led to higher error margins, indicating that integrated feature representations are critical for accurate forecasts.

\begin{table*}[htbp]
\caption{Performance metrics for predicting changes of ADAS-Cog and MMSE with 95\% confidence intervals (CI). The values in parentheses represent the lower and upper bounds of the 95\% CI. The metrics represent the predicted annual change (\(\Delta\)) for MMSE and ADASCOG. Note: This Characteristics feature set does not include the current values of MMSE and ADAS-Cog.}
\resizebox{\textwidth}{!}{
\begin{tabular}{llllll}
\toprule
 & \textbf{Metric} & \textbf{MAE\textsubscript{\(\Delta\)ADASCOG}} & \textbf{MAE\textsubscript{\(\Delta\)MMSE}} & \textbf{RMSE\textsubscript{\(\Delta\)ADASCOG}} & \textbf{RMSE\textsubscript{\(\Delta\)MMSE}} \\
Model & Feature Set &  &  &  &  \\
\midrule
\multirow[t]{13}{*}{LightGBM} & All Features & 6.21 (4.66, 7.84) & 2.61 (1.98, 3.27) & 6.20 (4.67, 7.97) & 2.63 (2.01, 3.27) \\
 & Baseline & 6.76 (4.96, 8.79) & 2.58 (1.90, 3.40) & 6.70 (5.07, 8.70) & 2.56 (1.86, 3.29) \\
 & Characteristics & \textbf{5.58 (4.11, 7.29)} & \textbf{2.55 (1.83, 3.42)} & \textbf{5.65 (4.13, 7.39)} & \textbf{2.55 (1.82, 3.38)} \\
 & Proportion Baseline & 6.08 (4.64, 7.71) & 2.66 (2.01, 3.39) & 6.13 (4.55, 7.82) & 2.66 (2.01, 3.35) \\
 & Random Word & 5.68 (4.28, 7.36) & 2.75 (2.13, 3.43) & 5.74 (4.26, 7.59) & 2.76 (2.13, 3.43) \\
 & State & 6.01 (4.49, 7.77) & 2.68 (2.04, 3.40) & 5.97 (4.37, 7.68) & 2.69 (2.05, 3.49) \\
\cline{1-6}
\multirow[t]{13}{*}{SVM} & All Features & 6.32 (4.84, 8.11) & 3.61 (3.05, 4.32) & 6.25 (4.74, 7.99) & 3.63 (3.02, 4.34) \\
 & Baseline & 6.22 (4.80, 7.97) & 3.61 (3.02, 4.35) & 6.25 (4.54, 8.02) & 3.62 (2.97, 4.32) \\
 & Characteristics & 6.27 (4.72, 8.04) & 3.60 (2.95, 4.29) & 6.33 (4.69, 8.01) & 3.60 (2.98, 4.35) \\\
 & Proportion Baseline & 6.24 (4.80, 7.93) & 3.60 (2.96, 4.33) & 6.27 (4.64, 7.87) & 3.60 (3.01, 4.33) \\
 & Random Word & 6.28 (4.64, 7.96) & 3.61 (2.97, 4.33) & 6.30 (4.65, 8.00) & 3.61 (2.98, 4.35) \\
 & State & 6.25 (4.82, 8.02) & 3.61 (2.97, 4.32) & 6.26 (4.60, 7.92) & 3.62 (2.97, 4.34) \\
\cline{1-6}
\multirow[t]{13}{*}{XGBoost} & All Features & 6.77 (5.17, 8.47) & 3.36 (2.51, 4.31) & 6.78 (5.18, 8.52) & 3.34 (2.49, 4.28) \\
 & Baseline & 6.15 (4.35, 8.08) & 2.91 (2.15, 3.75) & 6.10 (4.43, 8.02) & 2.92 (2.12, 3.78) \\
 & Characteristics & 7.87 (6.05, 9.92) & 3.41 (2.53, 4.42) & 7.83 (5.99, 9.73) & 3.41 (2.44, 4.41) \\
 & Proportion Baseline & 6.93 (5.27, 8.74) & 3.46 (2.67, 4.36) & 6.91 (5.27, 8.80) & 3.46 (2.66, 4.32) \\
 & Random Word & 7.18 (5.27, 9.17) & 2.83 (2.07, 3.66) & 7.15 (5.35, 9.19) & 2.83 (2.05, 3.72) \\
 & State & 7.24 (5.55, 9.10) & 3.31 (2.57, 4.07) & 7.26 (5.64, 9.11) & 3.31 (2.60, 4.08) \\
\cline{1-6}
\multirow[t]{13}{*}{LSTM} & Location Sequences& 27.78 (13.98, 41.57) & 27.23 (14.64, 39.82) & 35.95 (18.61, 47.31) & 34.63 (19.31, 45.00) \\

\bottomrule
\end{tabular}}
\label{change_nocurrent}
\end{table*}

The table presents the performance metrics for predicting annual changes in ADAS-Cog and MMSE (\(\Delta\)ADASCOG and \(\Delta\)MMSE) across various feature sets, with 95\% confidence intervals (CI). Notably, the clinical characteristics feature set used in this analysis excludes current values of MMSE and ADAS-Cog scores.

For the LightGBM model, the best predictive performance was achieved when using the Characteristics feature set alone, yielding an MAE\textsubscript{\(\Delta\)ADASCOG} of \textbf{5.58} (95\% CI: 4.11, 7.29) and an MAE\textsubscript{\(\Delta\)MMSE} of \textbf{2.55} (95\% CI: 1.83, 3.42). These results were the lowest across all LightGBM feature combinations, suggesting that the clinical characteristics features (excluding current MMSE and ADAS-Cog values) effectively capture predictive signals for cognitive decline.

In the XGBoost model, the Baseline + Characteristics feature set demonstrated competitive performance, with an MAE\textsubscript{\(\Delta\)ADASCOG} of 6.36 (95\% CI: 4.76, 8.34) and an MAE\textsubscript{\(\Delta\)MMSE} of 3.17 (95\% CI: 2.31, 4.12). However, using only the Characteristics feature set led to higher errors, with MAE values of 7.87 for ADAS-Cog and 3.41 for MMSE, highlighting the added value of baseline features in this context.

For the LSTM model, which utilized sequential location data, the performance was substantially lower compared to other models. The MAE\textsubscript{\(\Delta\)ADASCOG} was 27.78 (95\% CI: 13.98, 41.57), and the MAE\textsubscript{\(\Delta\)MMSE} was 27.23 (95\% CI: 14.64, 39.82). This suggests that while LSTM is adept at modeling sequential data, it may require more extensive feature engineering or larger datasets to achieve competitive performance in predicting cognitive decline.

As summarized in Table \ref{change_nocurrent}, excluding the current values of MMSE and ADAS-Cog slightly impacts predictive performance, but specific combinations of features such as Characteristics alone or Baseline + Characteristics still deliver robust results. These findings underscore the critical role of well-selected feature sets in improving the accuracy of cognitive decline predictions.

\begin{table*}[htbp]
\caption{Performance metrics for predicting current ADAS-Cog and MMSE with 95\% confidence intervals (CI). The values in parentheses represent the lower and upper bounds of the 95\% CI. }
\resizebox{\textwidth}{!}{
\begin{tabular}{llllll}
\toprule
 & \textbf{Metric} & \textbf{MAE\textsubscript{ADASCOG}} & \textbf{MAE\textsubscript{MMSE}} & \textbf{RMSE\textsubscript{ADASCOG}} & \textbf{RMSE\textsubscript{MMSE}} \\
Model & Feature Set &  &  &  &  \\
\midrule
\multirow[t]{13}{*}{LightGBM} & All Features & 12.96 (10.39, 15.48) & 4.90 (3.98, 5.82) & 12.99 (10.74, 15.49) & 4.92 (4.04, 5.78) \\
 & Baseline & 12.38 (10.33, 14.65) & 4.73 (3.92, 5.61) & 12.27 (10.05, 14.44) & 4.75 (3.94, 5.58) \\
 & Characteristics & \textbf{10.60 (8.45, 12.70)} & 4.27 (3.44, 5.14) & 10.64 (8.61, 12.83) & 4.28 (3.48, 5.20) \\
 & Proportion Baseline & 11.57 (9.37, 13.82) & 4.79 (3.96, 5.73) & 11.60 (9.40, 13.97) & 4.80 (3.99, 5.71) \\
 & Random Word & 10.77 (9.01, 12.67) & 4.40 (3.61, 5.23) & 10.76 (8.92, 12.53) & 4.35 (3.59, 5.17) \\
 & State & 10.64 (8.67, 12.70) & \textbf{4.24 (3.40, 5.14)} & \textbf{10.60 (8.60, 12.69)} & \textbf{4.24 (3.37, 5.10)} \\
\cline{1-6}
\multirow[t]{13}{*}{SVM} & All Features & 10.85 (8.41, 13.28) & 4.71 (3.79, 5.71) & 10.77 (8.37, 13.22) & 4.71 (3.84, 5.62) \\
 & Baseline & 10.76 (8.38, 13.28) & 4.51 (3.60, 5.45) & 10.66 (8.10, 13.13) & 4.48 (3.62, 5.36) \\
 & Characteristics & 10.89 (8.59, 13.56) & 4.44 (3.46, 5.42) & 10.90 (8.72, 13.32) & 4.46 (3.53, 5.45) \\
 & Proportion Baseline & 10.72 (8.47, 13.15) & 4.46 (3.63, 5.44) & 10.69 (8.29, 13.34) & 4.47 (3.61, 5.41) \\
 & Random Word & 10.77 (8.52, 13.25) & 4.40 (3.49, 5.41) & 10.79 (8.51, 13.30) & 4.39 (3.52, 5.32) \\
 & State & 10.62 (8.19, 13.08) & 4.42 (3.50, 5.38) & 10.72 (8.26, 13.12) & 4.39 (3.49, 5.33) \\
\cline{1-6}
\multirow[t]{13}{*}{XGBoost} & All Features & 11.84 (9.50, 14.12) & 4.86 (3.95, 5.82) & 11.82 (9.40, 14.32) & 4.89 (3.96, 5.81) \\
 & Baseline & 11.64 (9.00, 14.39) & 5.66 (4.51, 6.75) & 11.60 (8.89, 14.58) & 5.65 (4.62, 6.71) \\
 & Characteristics & 11.77 (8.70, 14.82) & 5.03 (3.76, 6.28) & 11.74 (9.07, 14.62) & 5.02 (3.74, 6.38) \\
 & Proportion Baseline & 11.57 (9.22, 14.09) & 5.29 (4.26, 6.33) & 11.63 (8.99, 14.26) & 5.28 (4.31, 6.32) \\
 & Random Word & 12.61 (9.81, 15.62) & 4.79 (3.77, 6.00) & 12.66 (9.97, 15.68) & 4.81 (3.78, 5.96) \\
 & State & 11.09 (8.53, 13.94) & 4.67 (3.56, 5.89) & 10.95 (8.38, 13.57) & 4.65 (3.62, 5.69) \\
\cline{1-6}
\multirow[t]{13}{*}{LSTM} & Location Sequences & 11.35 (8.56, 14.13) & 4.93 (3.94, 5.91) & 14.51 (11.66, 16.89) & 5.99 (4.76, 7.02) \\
\bottomrule
\end{tabular}}
\label{tab:current_table_2}
\end{table*}

Table~\ref{tab:current_table_2} summarizes the performance of various models (LightGBM, SVM, XGBoost, and LSTM) in predicting current ADAS-Cog and MMSE scores, evaluated through mean absolute error (MAE) and root mean square error (RMSE), with 95\% confidence intervals (CIs) provided in parentheses. The results highlight that feature sets significantly influence predictive performance. The LightGBM model achieved its best results using the "Characteristics" feature set, with MAE and RMSE values for ADAS-Cog of 10.60 (95\% CI: 8.45, 12.70) and 10.64 (95\% CI: 8.61, 12.83), and for MMSE of 4.27 (95\% CI: 3.44, 5.14) and 4.28 (95\% CI: 3.48, 5.20). Similarly, the XGBoost model performed optimally with the "State + Characteristics" feature set, achieving the lowest MAE for ADAS-Cog of 10.40 (95\% CI: 8.12, 12.59) and for MMSE of 4.13 (95\% CI: 3.21, 5.15), along with the lowest RMSE for ADAS-Cog of 10.44 (95\% CI: 8.38, 12.65) and for MMSE of 4.13 (95\% CI: 3.19, 5.07). The SVM model showed competitive performance with the "State" feature set, achieving MAE values of 10.62 (95\% CI: 8.19, 13.08) for ADAS-Cog and 4.42 (95\% CI: 3.50, 5.38) for MMSE, and RMSE values of 10.72 (95\% CI: 8.26, 13.12) for ADAS-Cog and 4.39 (95\% CI: 3.49, 5.33) for MMSE. In contrast, the LSTM model, which used location sequences as input, demonstrated relatively higher RMSE values of 14.51 (95\% CI: 11.66, 16.89) for ADAS-Cog and 5.99 (95\% CI: 4.76, 7.02) for MMSE, suggesting the need for further optimization. Overall, models utilizing feature sets enriched with "Characteristics" and "State + Characteristics" consistently outperformed those relying on baseline or random word features, and the inclusion of confidence intervals highlights the robustness of these results in predicting cognitive metrics.

\subsubsection{Ablation Study: Impact of Time-Series Length on Model Performance}

To investigate how the length of the time-series input affects the model's predictive performance, we conducted an ablation study by varying the time durations provided to the ridge model. This analysis specifically focuses on assessing the impact of input data collected over different periods: 7 days, 15 days, 30 days, 90 days, and 180 days. By analyzing these variations, we aim to understand the optimal time frame and seasonal changes for aggregating daily activity data to predict cognitive scores effectively.

\begin{figure*}[htbp]
    \centering
    \includegraphics[width=0.8\textwidth]{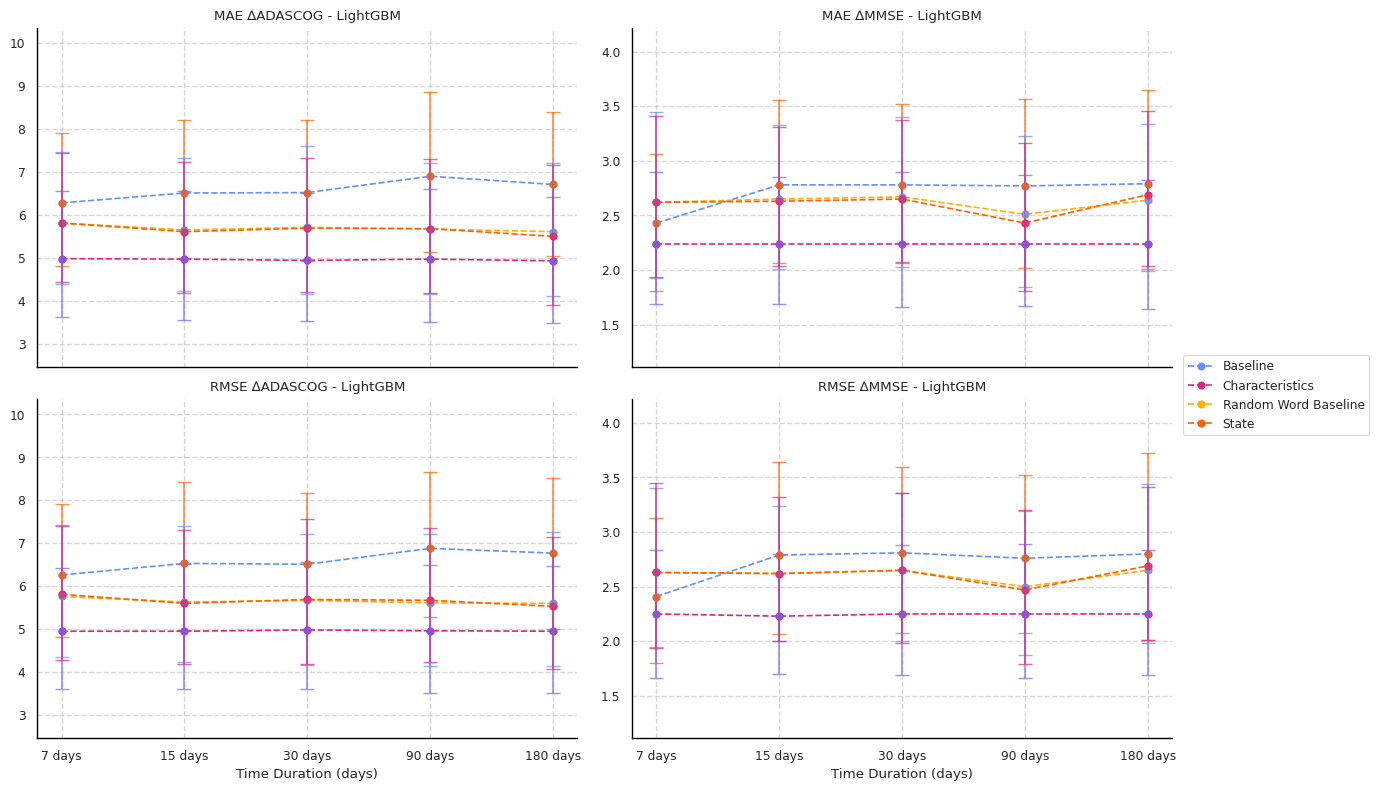}
    \caption{Prediction error for cognitive change (\( \Delta \)ADAS-Cog and \( \Delta \)MMSE). This figure shows the MAE and RMSE for cognitive change predictions using varying time-series lengths, including baseline, characteristics, and state feature sets. }
    \label{fig:metrics_data_change_ridge}
\end{figure*}

As shown in Figure \ref{fig:metrics_data_change_ridge}, which visualizes the prediction errors across varying time durations, we observed that the model generally achieved lower prediction errors when using longer time-series inputs (30 to 180 days). Specifically, for both MAE and RMSE metrics, the model's performance slightly degraded as the time duration less than 30 days. This trend was consistent across all metrics, suggesting that more participant movement data holds greater predictive value for cognitive change.

\begin{figure*}[htbp]
    \centering
    \includegraphics[width=0.8\textwidth]{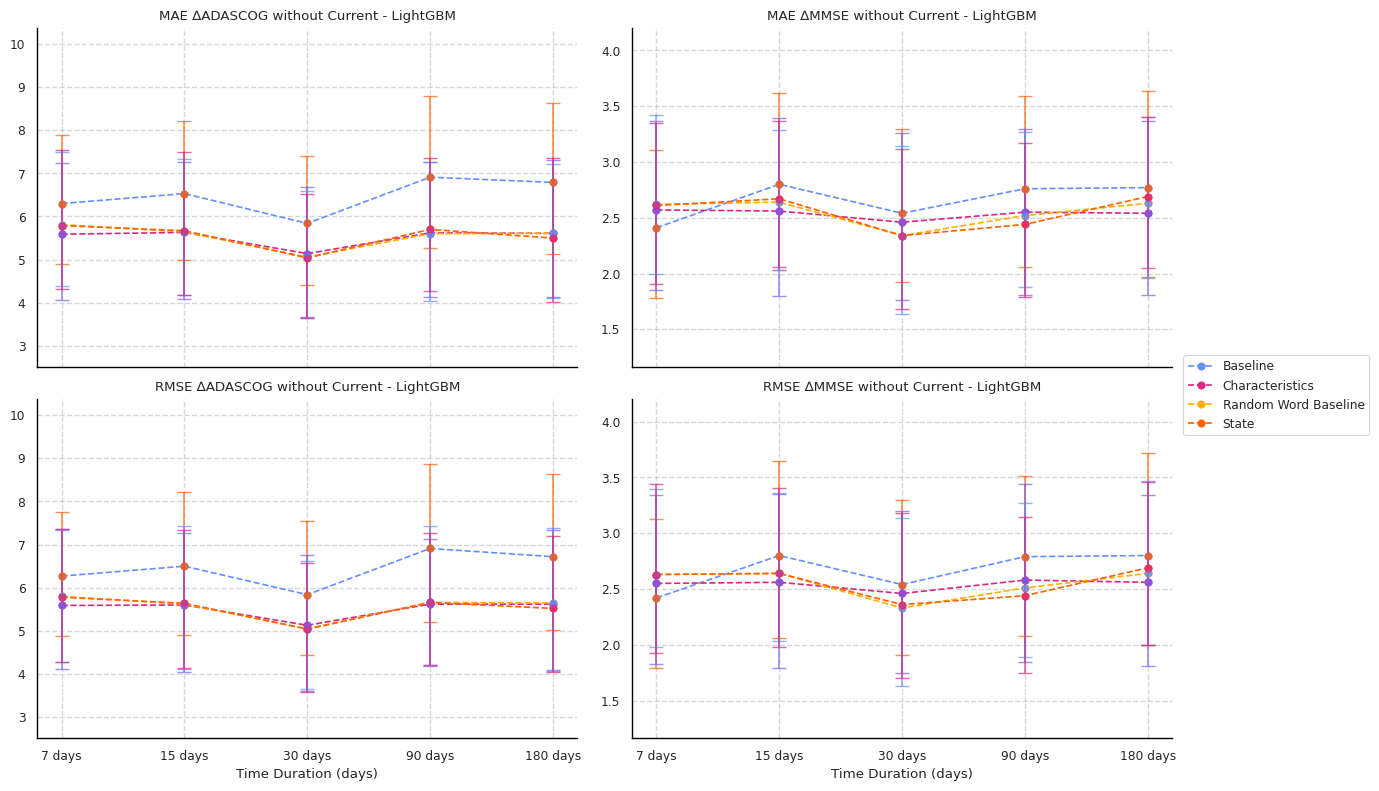}
    \caption{Prediction error for cognitive change (\( \Delta \)ADAS-Cog and \( \Delta \)MMSE) without current cognitive score in feature set. This figure illustrates the impact on prediction errors (MAE and RMSE) for cognitive change when excluding the current cognitive score from the feature set across different time durations.}
    \label{fig:metrics_data_change_no_current_ridge}
\end{figure*}

When excluding the current ADAS-Cog and MMSE values from the feature set, as depicted in Figure \ref{fig:metrics_data_change_no_current_ridge}, the error values showed a similar pattern. The absence of current scores slightly increased the MAE and RMSE, reinforcing the importance of longitudinal behavioral data for accurate predictions.

\subsubsection{Prediction of Absolute Cognitive Scores (ADASCOG and MMSE)}

\begin{figure*}[htbp]
    \centering
    \includegraphics[width=0.8\textwidth]{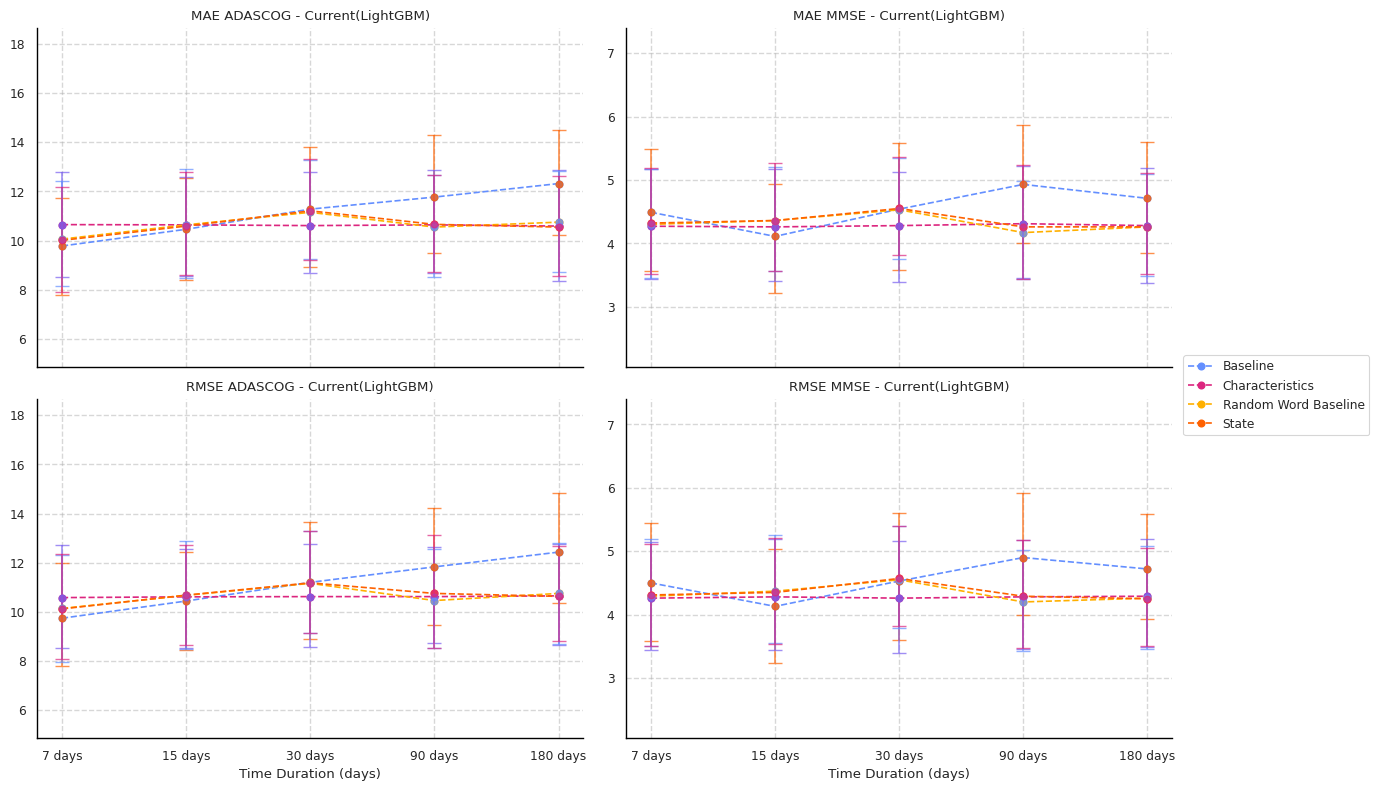}
    \caption{Prediction error for current cognitive scores (ADAS-Cog and MMSE). This figure depicts the MAE and RMSE for absolute cognitive score predictions, showing the performance stability across various time-series lengths with baseline, characteristics, and state feature sets.}
    \label{fig:metrics_data_current_ridge}
\end{figure*}

In contrast, Figure \ref{fig:metrics_data_current_ridge}, which represents the performance metrics for predicting absolute ADAS-Cog and MMSE scores. When predicting current value of cognitive scores, the MAE and RMSE remained relatively stable across time durations, with only minor decreases in error as the time frame extended to 180 days. This stability suggests that longer time-series data can be valuable for its cognitive status information.

\subsubsection{State Feature Performance}
For this ablation study, we observed that as the input time span increases, the performance gap between the state model and the baseline model widens. This finding further supports that the state feature—derived from our deep analysis of patient mobility patterns—captures significant information about patient behavioral patterns and cognitive abilities. Additionally, with longer input time series, the state model demonstrates greater robustness in predicting cognitive function compared to the baseline, showcasing its capacity to capture long-term temporal information, particularly when the time series is exceptionally extended.

Moreover, in nearly all experiments, the performance of the state feature, which only depicts the movement of patients, is comparable to that of the characteristic feature. In the case of extended time series, the state feature even outperforms the characteristic feature in prediction accuracy. This indicates that the state feature we generate, potentially offering deep insight in clinical diagnostic results, may be a effective predictor of cognitive status.

\subsection{Clinical Challenges}
From a clinical perspective, aligning the outputs of AI models with actionable insights for healthcare providers is a key challenge. The proposed low-rank state representations must correlate strongly with meaningful clinical metrics such as MMSE and ADAS-Cog scores to support diagnosis and intervention planning. However, the limited size and scope of our dataset, which includes only 134 participants with complete data from 50 patients, presents a significant limitation. This small sample size may not fully capture the diversity of movement and behavioral patterns observed in the wider population of individuals living with dementia. As a result, the model's generalizability to other settings or populations could be compromised. Additionally, the variability in patient behavior due to external factors, such as seasonal changes or caregiver interactions, necessitates models that can differentiate between pathological changes and normal variations. Integrating AI-driven insights into existing clinical workflows without increasing the burden on healthcare providers remains a critical consideration to ensure the practical utility of these systems.

Addressing these challenges requires an interdisciplinary approach, incorporating expertise from AI researchers, clinicians, ethicists, and regulatory bodies, to develop robust, equitable, and clinically impactful healthcare monitoring solutions. Expanding the dataset to include a more diverse and representative sample of patients will also be essential for enhancing the model’s reliability and applicability.

\end{document}